\documentclass[lettersize,journal]{IEEEtran}
\usepackage{amsmath,amsfonts}
\usepackage{algorithmic}
\usepackage{array}
\usepackage[caption=false,font=normalsize,labelfont=normalsize,textfont=normalsize]{subfig}
\usepackage{textcomp}
\usepackage{stfloats}
\usepackage{url}
\usepackage{verbatim}
\usepackage{graphicx}
\def\BibTeX{{\rm B\kern-.05em{\sc i\kern-.025em b}\kern-.08em
    T\kern-.1667em\lower.7ex\hbox{E}\kern-.125emX}}
\usepackage{balance}

\usepackage{amssymb} % \intercal
\usepackage{multirow} % \multirows
\usepackage{cancel} % \cancelto
\usepackage{xcolor}
\usepackage{hyperref}

\newcommand{\etal}{\textit{et al}.}

\newcommand{\twist}[2]{{\mathcal V}^\text{#1}_\text{#2}}
\newcommand{\wrench}[2]{{\mathcal F}^\text{#1}_\text{#2}}
\newcommand{\disp}[2]{{\mathcal X}^\text{#1}_\text{#2}}
\newcommand{\AdT}[1]{[\text{Ad}_{{\bf T}_{#1}}]}
\newcommand{\R}[1]{{\bf R}_{#1}}
\newcommand{\p}[2]{{\bf p}^\text{#1}_\text{#2}}
\newcommand{\stiffness}[1]{{\bf K}^{#1}_\text{spr}}
\newcommand{\W}[2]{{\bf W}^\text{#1}_\text{#2}}
\newcommand{\ww}[2]{{\boldsymbol \omega}^\text{#1}_\text{#2}}
\newcommand{\vv}[2]{{\bf v}^\text{#1}_\text{#2}}

\newcommand{\curvature}[1]{{\bf K}_\text{cur{#1}}}
\newcommand{\torsion}[1]{{\bf T}_\text{tor{#1}}}
\newcommand{\metric}[1]{{\bf M}_\text{met{#1}}}
\newcommand{\real}{\mathbb{R}}

\begin{document}

\title{Compliant In-hand Rolling Manipulation Using Tactile Sensing}
    
\author{Huan Weng, Yifei Chen, and Kevin M. Lynch,~\IEEEmembership{Fellow,~IEEE}

\thanks{The authors are affiliated with the Center for Robotics and Biosystems and the Department of Mechanical Engineering, Northwestern University, Evanston, IL 60208 USA.
Huan Weng is currently at Sudo AI, Shanghai, China, and Kevin Lynch is affiliated with the Northwestern Institute on Complex Systems (NICO). 
{Email: \tt\footnotesize huanweng@u.northwestern.edu}, 
{\tt\footnotesize yifeichen2026@u.northwestern.edu},
{\tt\footnotesize kmlynch@northwestern.edu}.}
\thanks{This work was funded in part by the US NSF HAND Engineering Research Center, 610-4734006-60069218.}
}

\markboth{To appear, \textit{IEEE Transactions on Robotics}}
{Weng \MakeLowercase{\textit{et al.}}: Compliant In-hand Rolling}

\maketitle

\begin{abstract}
We investigate in-hand rolling manipulation using a multifingered robot hand, where each finger is compliant and equipped with a tactile fingertip providing contact location and wrench information.  
We derive the equations of motion and inverse mechanics for compliant quasistatic in-hand rolling manipulation,  and we formulate a fingertip rolling manipulation controller for multiple fingers to achieve a desired object twist within a grasp. The approach is validated in simulation, and the controller is tested on an experimental robot system.
\end{abstract}


\section{Introduction}
\IEEEPARstart{P}{erhaps} the biggest challenge in creating useful humanoid robots is the ``hands problem'': achieving dexterity
approaching or exceeding human manual dexterity~\cite{HumDex2025,WSJ2025}. 

\begin{figure}
\centering
\includegraphics[width = 3in]{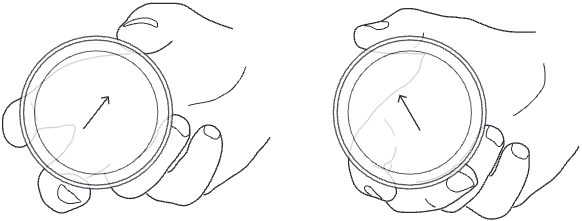}
\caption{Most in-hand manipulation requires rolling, sliding, and/or spinning at finger contact patches, as illustrated in this example of three-fingered in-hand twisting of the lid of a jar.}
\label{fig:human_demo}
\end{figure}

\begin{figure}
\centering
\includegraphics[width = 1.7in]{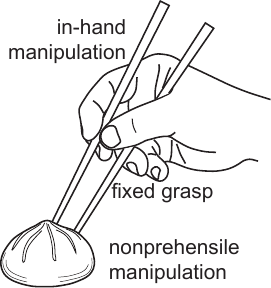}
\caption{Three major categories of dexterous manipulation in a common example of tool use. The lower chopstick is held in a fixed grasp, the upper chopstick undergoes in-hand manipulation, and the chopsticks push the steamed bun (nonprehensile manipulation). Figure adapted from~\cite{shilati2026benchmarking} with permission.}
\label{fig:chopsticks-3-types}
\end{figure}

Dexterous manipulation by hands can be divided into two broad categories: nonprehensile manipulation (such as pushing) and grasping manipulation. Grasping manipulation can be further divided into fixed grasps, where the manipulated object remains fixed relative to a hand frame and all manipulation is performed by the arm, and in-hand manipulation, where fingers control the state of the object relative to the hand frame, as when twisting the lid of a jar (Figure~\ref{fig:human_demo}). These manipulation types also apply to manipulation by multiple hands, or of multiple objects. Different manipulation types can occur simultaneously, even in a single hand; for example, during typical use of chopsticks, the lower chopstick may be held in a fixed grasp while the upper chopstick undergoes in-hand manipulation (Figure~\ref{fig:chopsticks-3-types}).

A simple kinematic analysis of a rigid body within a multifingered grasp shows that, unless each grasping finger has at least six degrees of freedom, achieving full in-hand mobility of the grasped object relative to the palm requires relative motion at the grasping contacts (Figure~\ref{fig:human_demo}). Examples of relative motion at a contact include sliding in the contact tangent plane, which provides two degrees of freedom of mobility at the contact; spinning about the contact normal, which provides one degree of freedom of mobility; no-spin ``pure rolling''~\cite{Woodruff2023}, i.e., rotation about two axes in the contact tangent plane but not about the contact normal axis, which provides two degrees of freedom of mobility; rolling, which is a combination of spin and pure rolling; and general roll-slide, which provides five degrees of freedom of mobility. 

\begin{figure}
\centering
\includegraphics[height=2.6in]{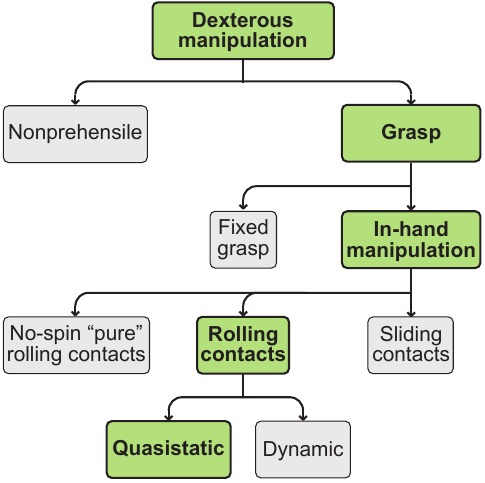}
\caption{A classification of types of dexterous manipulation. This paper focuses on quasistatic in-hand rolling manipulation of rigid bodies.}
\label{fig:classification}
\end{figure}

The focus of this paper is controlling in-hand manipulation of rigid bodies, specifically using three-degree-of-freedom rolling at fingertips (Figure~\ref{fig:classification}). Control of relative motion at the fingertips is facilitated by tactile sensing, and desirable features of tactile fingertips include
\begin{itemize}
    \item contact location sensing;
    \item contact wrench sensing;
    \item a smooth, strictly convex shape ensuring a positive-definite relative curvature form at point contacts with convex objects, enabling rolling; and
    \item passive compliance, to allow safe and stable grasping of rigid objects while avoiding mechanical overconstraint.
\end{itemize}

\begin{figure}
\centering
\includegraphics{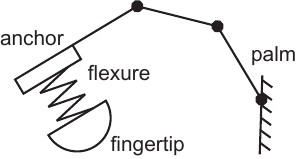}
\caption{Simplified model of a finger. The ``anchor'' is driven by high-impedance, effectively motion-controlled, finger joints, typical of the highly-geared actuators used in many robot hands. The fingertip is mounted to the anchor by a 6-dof flexure, providing passive compliance and the possibility of force control for safe manipulation of rigid objects.  This model can be shown to be equivalent to models where the compliance is distributed to the finger joints or links, or implemented via active stiffness control.}
\label{fig:finger_motion}
\end{figure}

To model passive compliance, we use the anchor-fingertip model of Figure~\ref{fig:finger_motion}. The controls are joint velocities of the fingers, a model consistent with the high impedance of many robot hand designs that rely on high gear ratio transmissions. These joint velocities directly generate a twist at the ``anchor'' frame at the distal link of each finger. The anchor is connected to the fingertip via a 6-dof flexure, providing passive compliance for safe manipulation of rigid objects. In this model, all passive compliance is concentrated at the flexure, but a mathematical equivalence can be established to fingers with distributed compliance or compliance at the joints, e.g., joints actuated by series-elastic actuators or compliant tendons. At low bandwidths, the compliance of the anchor-fingertip model can be emulated by active stiffness control~\cite{shi2020hand}.

In-hand manipulation with compliant fingers results from the controlled motion of the anchors, the compliance, the shapes of the object and the fingertips, and friction at the contacts (e.g., Figure~\ref{fig:sim}). To derive a mathematical model of in-hand rolling manipulation, we make the following assumptions: (1) the hand manipulates a single rigid object in free space; (2) contacts are confined to fingertips, and contact patches are small and can be modeled as points; (3) contacts are governed by dry Coulomb friction; and (4) inertial forces are negligible (quasistatic mechanics).

\begin{figure}
\centering
\includegraphics[width=3.2in]{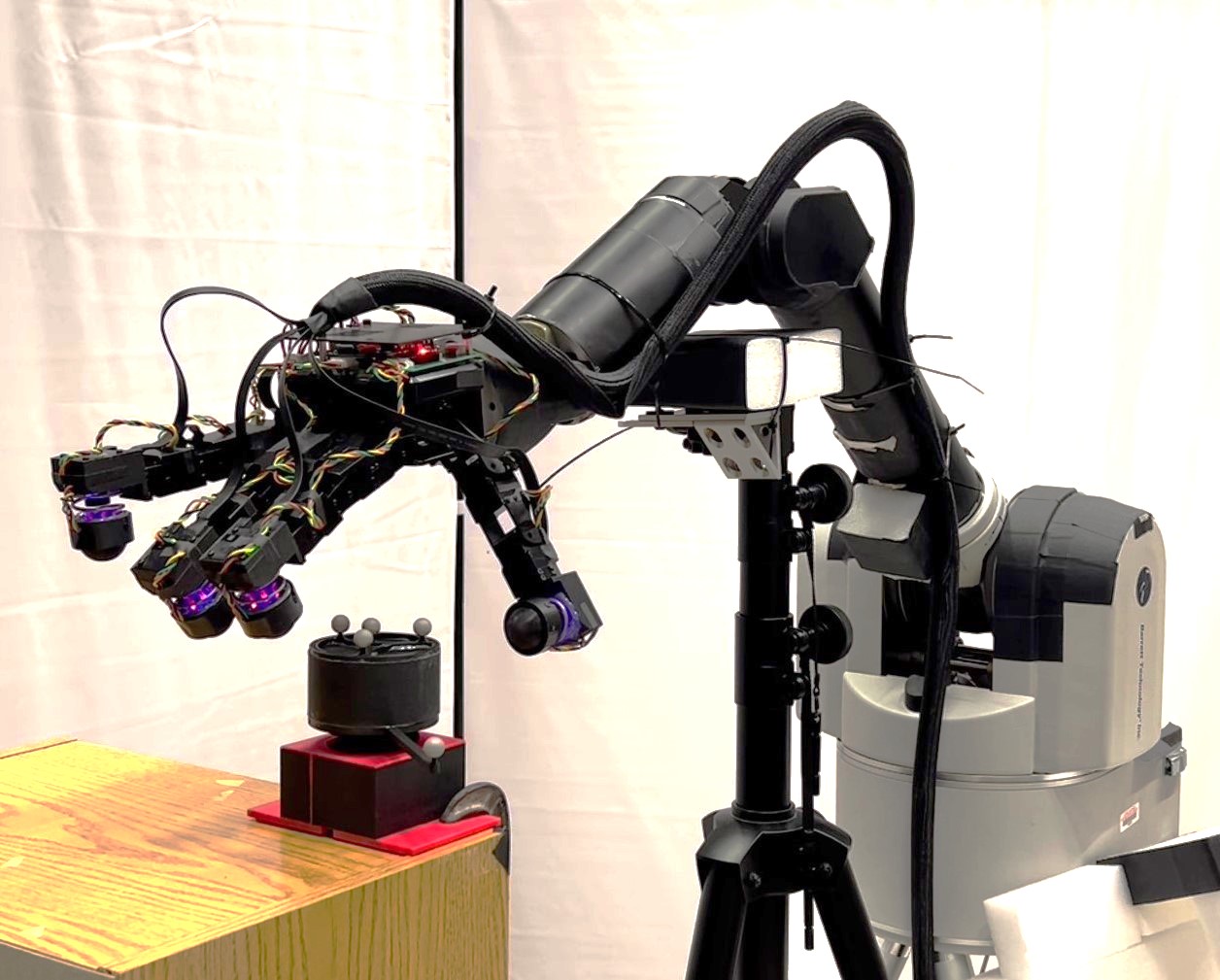}
\caption{The Barrett WAM robot arm and Allegro hand equipped with Visiflex tactile sensors.} 
\label{fig:overview}
\end{figure}

For experimental implementation, we use an Allegro robot hand with each finger equipped with a Visiflex tactile fingertip~\cite{fernandez2021visiflex}, as shown in Figure~\ref{fig:overview}. The Visiflex is an optical tactile sensor with a camera in the anchor, a spring steel flexure, and a hemispherical light-infused waveguide fingertip. The Visiflex is designed for (1) well-characterized compliance due to the flexure, (2) optical contact location sensing, and (3) optical 6-dof contact wrench sensing, using the imaged displacement of the fingertip and a model of the flexure's stiffness. The fingertip's hemispherical shape facilitates rolling contact, which is not possible for flat tactile fingertips manipulating objects with flat sides. 
Real-time wrench and contact location feedback, coupled with the control strategy explored in this paper, can enable robust execution of in-hand rolling manipulation, such as twisting the lid in Figure~\ref{fig:human_demo}.

The contributions of this paper include
\begin{itemize}
\item derivation of the quasistatic contact mechanics of a robot hand with multiple compliant fingertips in rolling contact with a grasped object; 
\item a control algorithm that uses these mechanics and tactile sensing to control the motion of a grasped object to follow a pre-planned trajectory; and 
\item simulation and preliminary experimental validation of the approach. 
\end{itemize}

This paper does not provide a complete solution to planning and control of in-hand rolling manipulation. In particular, it does not address the problems of (1) grasp planning, (2) control for grasp acquisition, (3) planning of a feasible in-hand rolling trajectories, or (4) breaking and re-establishing contact while maintaining a grasp (finger gaiting) for large object motions. The focus is differential mechanics and control in an established grasp. 

This paper adopts a model-based approach, which clarifies the role of compliance and geometry in rolling manipulation of rigid objects but requires accurate models of the hand and object for practical implementation. For cases where accurate models are not available and a data-driven approach is warranted, an understanding of the compliant rolling mechanics, provided by the model-based approach, can be used to (1) focus robot experiments or (2) extract more information from experiments to achieve more efficient robot learning (e.g., Section~\ref{ssec:model-assisted}).

The remainder of the paper is organized as follows.
Section~\ref{cha:related} presents related work. Section~\ref{cha:model_preli} introduces mathematical preliminaries for contact modeling. Section~\ref{sec:multi_fingers} derives the contact mechanics of multiple fingers rolling an object within a compliant grasp and proposes a controller for in-hand rolling manipulation. Section~\ref{cha:experiments} presents preliminary experiments validating the approach. Section~\ref{cha:conclusion} concludes and presents avenues for future research. 

%%%%%%%%%%%%%%%%%%%%%%%%%%%%%%%%%
%%%%%%%%%%%%%%%%%%%%%%%%%%%%%%%%%

\section{Related work} \label{cha:related}
\subsection{Tactile sensing for dexterous manipulation}
\noindent For decades, various types of tactile sensors, along with corresponding data processing methods, have been developed to detect fingertip slip, including but not limited to \cite{holweg1996slip, melchiorri2000slip, james2018slip, james2020slip, dong2019maintaining}. Lynch~\etal\ used active tactile sensing to estimate an object's center of mass by pushing it~\cite{lynch1992manipulation}. Sundaralingam and Hermans estimated objects' friction and inertial parameters using vision and tactile fingertip sensing. The parameters were estimated by standard numerical optimization based on a factor graph representing the dynamics between the fingertips and the object \cite{sundaralingam2021hand}. Suresh \etal\ developed NeuralFeels, combining vision and touch sensing to perform shape and pose estimation during in-hand manipulation \cite{suresh2024neuralfeels}. Di \etal\ leveraged optical fiber bundles to reduce the form factor of their tactile sensor, DIGIT Pinki, and utilized it in teleoperated digital palpation \cite{di2024using}.

Tactile feedback can also guide reactive manipulation control strategies. Rodriguez emphasized the use of tactile feedback to regulate contact modes in manipulation, thereby simplifying the dynamics of frictional contact \cite{rodriguez2021unstable}. 
Rodriguez \etal\ also developed a series of tactile sensors, including the GelSlim \cite{donlon2018gelslim, ma2019dense, dong2019maintaining} variation of GelSight \cite{li2014localization, yuan2017gelsight}, and used them for dexterous manipulation tasks such as local grasp adjustments \cite{hogan2018tactile}, cable manipulation \cite{she2021cable}, and extrinsic contact sensing \cite{ma2021extrinsic}. Maria \etal\ developed force/tactile sensors \cite{de2012force, d2011silicone} and different model-based strategies to avoid slippage \cite{cirillo2017control, de2013slipping} and achieve slipping control \cite{costanzo2018slipping, costanzo2019two}. Similar to our approach, Dollar \etal\ integrated contact sensing on a well-tuned compliant hand to improve grasp success rate \cite{dollar2010contact}. Hoof \etal\ used reinforcement learning to acquire in-hand manipulation skills, avoiding the modeling complexity from compliant hands and tactile sensors \cite{van2015learning}.  
Khadivar and Billard developed a control framework for robust grasp and in-hand manipulation based on coupled dynamical systems \cite{khadivar2023adaptive}. The controller relied on tactile sensors whose information was used to estimate the contact frames for all fingertips through a learned model. Zhao \etal\ proposed a novel tactile-informed manipulation method, Tac-Man, that can robustly maintain stable contact during manipulation of articulated objects but requires no prior-based modeling \cite{zhao2024tac}. Yuan \etal\ proposed Robot Synesthesia, a visuotactile approach to perform in-hand rotation of a single or multiple objects \cite{yuan2024robot}. Their policy was trained in simulation but can be effectively transferred to the real world.

Model-based, data-driven, and combined methods have been explored for processing different types of tactile information, as reviewed in \cite{howe1993tactile, tegin2005tactile, li2020review}.

\subsection{Compliant grasp}
Early work on grasping using compliant multi-fingered hands can be traced back to \cite{hanafusa1977stable}, when Hanafusa and Asada proposed a control strategy for stable prehension. Stable prehension is defined by force balance and minimum potential energy. The existence of stable grasps for different geometries was analyzed across various hand-finger configurations. Based on this model, Baker \etal\ studied the stability of different grips on planar polygons by investigating the local minima of the potential energy function \cite{baker1985stable}. 

Building on this work, Cutkosky and Kao analyzed the dependence of grasp compliance on grasp configuration, contact conditions, and the mechanical properties of fingers, using finger joint stiffnesses to achieve a desired grasp compliance~\cite{cutkosky1989computing}. 
Bruyninckx \etal\ developed a differential geometric framework to measure the stability of compliant grasps based on the generalized eigen-decomposition of the grasp stiffness matrix \cite{bruyninckx1998generalized}. Malvezzi and Prattichizzo proposed a grasp compliance model based both on the structural compliance of the fingers and the gains of impedance controllers \cite{malvezzi2013evaluation}.

The robustness of compliant grasps to object pose uncertainties has been addressed in~\cite{chen2015adaptive}, where compliance is modeled as controlled virtual springs between the fingertips, and in~\cite{pozzi2016grasp}, which proposes an index measuring the robustness of compliant grasps. 

Zhou \etal\ demonstrated that finger compliance can be controlled using soft hydraulic actuation, and that this finger compliance can be used to perform in-hand manipulation~\cite{zhou2024dexterous}. 

\subsection{In-hand manipulation}
\noindent 
Dafle \etal\ defined ``extrinsic dexterity'' as using external forces to enhance the dexterity of a robot hand in regrasp operations. A grasp graph is generated to transit between different grasps for complicated tasks \cite{dafle2014extrinsic}. For rolling, Montana derived the first-order kinematics of contact between two rigid bodies, including the contact equations for pure rolling \cite{montana1988kinematics}. Cole \etal\ developed the kinematics, prehensility, dynamics, and control of in-hand manipulation of objects with arbitrary shapes by multiple rolling fingers \cite{cole1988kinematics}. They also presented a dynamic control law for sliding regrasp in planar cases \cite{cole1992dynamic}. 
Ward-Cherrier \etal\ achieved model-free active tactile manipulation using a tactile gripper. Cylinders are reoriented by rolling between gripper pads \cite{ward2017model}. Spiers \etal\ developed robotic fingers with variable friction surfaces, which meet different friction needs to achieve in-hand rolling and sliding, and thus can enhance hand dexterity \cite{spiers2018variable}. Lee \etal\ developed an object rolling controller based on the optimization of the trajectory of the finger movement while ensuring that contact points move along the fingers. The introduction of a force controller with compliance further improved the accuracy and robustness of the motion \cite{lee2025trajectory}. 

Building on trajectory optimization, recent frameworks have achieved large-range manipulation through geometry-free approaches \cite{rgmc2024geometryfree} and enabled complex contact sequences via hierarchical exploration \cite{hidex2025}. 
Machine learning has also been utilized to achieve in-hand dexterity~\cite{levine2016end, andrychowicz2020learning, gualtieri2020learning,rostel2023estimator,pitz2024learning}.
To bridge the sim-to-real gap, recent learning paradigms have introduced joint-wise neural dynamics modeling for robust in-the-air rotation \cite{liu2025dexndm} and hierarchical policies leveraging pre-trained skills to reorient complex objects \cite{qi2025simplecomplexskillscase}.

Hou \etal\ developed a strategy to achieve planar dynamic pivoting, a unique form of sliding regrasp, through wrist swing motion and grip force regulation. The strategy is robust to uncertainties in both friction and grasp position \cite{hou2020robust}. Dafle \etal\ generalized the construction of the motion cone for a broad set of planar manipulation tasks. Using this representation, they showed that regrasp could be achieved by a sequence of object pushes against external constraints when the gripper slides on the object \cite{chavan2020planar}. Shi \etal\ implemented passively compliant fingers to achieve sliding regrasp by taking advantage of external forces such as those from object inertial load \cite{shi2017dynamic} and contact between the object and a rigid environment \cite{shi2020hand}. 

To our knowledge, our work is the first to (1) explicitly model passive compliance in simultaneous quasistatic rolling at multiple smooth contacts and (2) use that model to achieve in-hand rolling manipulation. In most previous work, compliance is necessarily present but not explicitly acknowledged or modeled, limiting the potential transferability to tasks or hardware with different or unknown compliance properties.

\subsection{Physics-informed learning of in-hand manipulation}
\label{ssec:model-assisted}
\noindent 
Replicating the success of foundation models from natural language and computer vision in robotics is fundamentally obstructed by a ``100,000-year data gap''---the orders-of-magnitude disparity between internet-scale linguistic data and the scarcity of physical interaction data \cite{goldberg2025gofe}. To bridge this gap, recent research leverages physical priors as structural inductive biases to avoid unstructured end-to-end mapping and ensure physical consistency \cite{brooks2025dexterity, zhu2020overview}.

Recent works explicitly embed analytical priors into the learning loop via differentiable simulation or neural feature extraction. Differentiable simulators have been developed to propagate gradients through contact events, facilitating system identification and tactile policy transfer \cite{yang2025differentiable, si2024difftactile}. For instance, Hu \etal\ \cite{HU2025104904} guided reinforcement learning (RL) policies using contact center positions extracted from tactile sensors to achieve precise rolling of cylindrical objects. Notably, Lee and Fazeli \cite{lee2025vitascope} developed ViTaSCOPE, employing physics-constrained neural implicit representations to fuse vision and tactile feedback for contact field reconstruction.

Stability-aware constraints have also been integrated directly into the RL objective to ensure mechanical consistency. Recent approaches explicitly embedded force-closure metrics \cite{chen2025inhand, zurbrugg2025graspqp} or control barrier functions \cite{yang2025cbfrl} into the reward structure or policy optimization to prioritize stable grasp configurations. Such methods effectively focus robot experiments within physically valid state-space regions.

Where analytical models fall short due to unmodeled interactions, residual physics learning combines nominal analytical controllers with learned residual policies \cite{shirai2025pivoting, Abhishek2026residual}. For example, the Grasp-to-Act framework utilizes physics-based optimization to guide RL-driven adaptation \cite{gupta2025grasp}. 

Similarly, R{\"o}stel \etal\ leveraged RL critics to score initial grasps that promote specific in-hand manipulation goals \cite{Roestel_2025}. Physics-informed world models, such as RoboScape \cite{shang2025roboscape} and PIN-WM \cite{li2025pinwm}, extend this synergy by constraining latent dynamics with geometric consistency and rigid-body parameters for physically consistent policy refinement.

%%%%%%%%%%%%%%%%%%%%%%%%%%%%%%%%%
%%%%%%%%%%%%%%%%%%%%%%%%%%%%%%%%%

\section{Contact model and preliminaries} \label{cha:model_preli}
\subsection{Contact model}
\begin{figure}[!t]
\centering
\includegraphics[width= 3.4in]{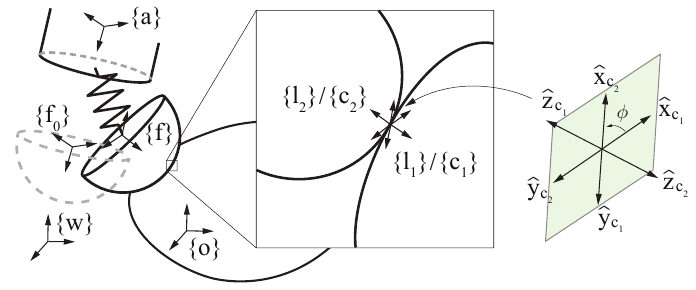}
\caption{A fingertip displaced from the flexure's relaxed position and making contact with an object, with defined frames. The fingertip's pose when the flexure is relaxed is drawn in grey. All the frames at the contact point have their $\hat{z}$-axes normal to the tangent plane. $\phi$ is the rotation angle about the $\hat{z}$-axis of frame \{l$_1$\}/\{c$_1$\} from the $\hat{x}$-axis of \{l$_1$\}/\{c$_1$\} to that of \{l$_2$\}/\{c$_2$\}.}
\label{fig:frames}
\end{figure}

\begin{figure}[!t]
\centering
\includegraphics[width= 3.4in]{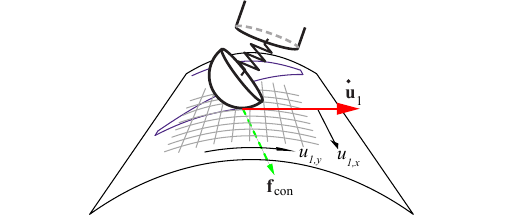}
\caption{A compliant fingertip moving on the object surface while applying the contact force ${\bf f}_\text{con}$. A local surface patch and a coordinate frame are defined with the origin at the contact point and two axes $u_{1,x}$ and $u_{1,y}$ orthogonal to each other. The contact point velocity can be represented in that frame as $\dot{\bf u}_1$.}
\label{fig:local_patch}
\end{figure}

Figure \ref{fig:frames} illustrates a moving compliant fingertip making a point contact with a moving object.
A local surface patch and coordinates with orthogonal axes $u_{1,x}$ and $u_{1,y}$ are defined at the contact point on the object to represent the contact velocity $\dot{\bf u}_1$ along the object surface \cite{montana1988kinematics}, as shown in Figure~\ref{fig:local_patch}.
Using the definitions in \cite{montana1988kinematics} and the finger spring compliance model in \cite{shi2020hand}, we define frames as listed in Table~\ref{table:frames}, and an extra index $i = 1\ldots n$ is used in subscripts to indicate the particular finger.
\begin{table}
\renewcommand{\arraystretch}{1.5}
\centering
\caption{Frame definitions}
\begin{tabular}{p{0.4in}p{2.7in}}
\hline
Frame & Definition\\
\hline
\{w\} & World frame.\\
\{o\} & Object frame, attached to the object.\\
\{a\} & Fingertip anchor frame connected to the robot hand palm through finger links (phalanges) and joints.\\
\{f\} & Fingertip frame, attached to the hemispherical fingertip, connected with \{a\} through the flexure.\\
\{f$_0$\} & Fingertip rest frame, attached to \{a\} but coincident with \{f\} when the flexure is at rest.\\
\{l$_1$\},\{l$_2$\} & Stationary contact frames at the current contact points, attached to \{o\} and \{f\}, respectively, each with its $\hat{z}$-axis pointing outwards and $\hat{x}$, $\hat{y}$-axes tangential to the local coordinate system of the surface patch.\\
\{c$_1$\},\{c$_2$\} & Moving contact frames at the contact points, attached to the moving contact on the object and fingertip and simultaneously coincident with \{l$_1$\} and \{l$_2$\}, respectively.\\ 
\hline
\end{tabular}
\label{table:frames}
\end{table}

\subsection{Notation}
Scalars are written in italic lower case. Vectors and matrices are written as bold letters, except for poses represented by exponential coordinates $\disp{}{} \in \real^6$, twists $\twist{}{} = (\boldsymbol{\omega}, {\bf v}) \in \real^6$, wrenches $\wrench{}{} =({\bf m},{\bf f}) \in \real^6$, and the gravity vector ${\mathfrak g}\in\real^3$. All variables are expressed in the world frame \{w\} unless noted otherwise in the superscripts, and other descriptions are noted in the subscripts. For example, $\twist{}{wo}$ indicates the twist of the object frame relative to the world frame, expressed in the world frame, and $\twist{o}{wa,2}$ indicates the twist of the anchor frame of finger~2 relative to the world frame, expressed in the object frame. Table~\ref{table:notation} summarizes most of the notation, drawn primarily from~\cite{montana1988kinematics} and \cite{lynch2017modern}.
\begin{table}
\renewcommand{\arraystretch}{1.5}
\centering
\caption{Notation from ~\cite{montana1988kinematics,lynch2017modern}}
\begin{tabular}{p{0.25in}p{2.85in}}
\hline
Variable & Description\\
\hline
$\p{}{ij}$ & Position vector pointing from the origin of frame \{i\} to that of \{j\}, $\p{}{ij}\in{\mathbb R}^3$.\\
$\R{\text{ij}}$ & Rotation matrix from frame \{i\} to \{j\}, $\R{ij}\in SO(3)$.\\
${\bf T}_\text{ij}$ & Transformation matrix from frame \{i\} to \{j\}, ${\bf T}_\text{ij}:=\begin{bmatrix}\R{\text{ij}}&\p{i}{ij}\\{\bf 0}&1\end{bmatrix}\in SE(3)$.\\
$\AdT{\text{ij}}$ & Adjoint representation of ${\bf T}_\text{ij}$, $\AdT{\text{ij}}:=\begin{bmatrix}\R{\text{ij}}&{\bf 0}\\{[\p{i}{ij}]}\R{\text{ij}}&\R{\text{ij}}\end{bmatrix}$.\\
$\disp{}{ij}$ & Exponential coordinates $\disp{}{ij}\in{\mathbb R}^6$ of ${\bf T}_\text{ij}$, ${\bf T}_\text{ij}=\text{exp}([\disp{}{ij}])$.\\
$\twist{}{ij}$ & Twist of frame \{j\} relative to \{i\}, $\twist{}{ij}:=\begin{bmatrix}\ww{}{ij}\\\vv{}{ij}\end{bmatrix}\in{\mathbb R}^6$.\\
$\wrench{}{}$ & Wrench, including moment and force, $\wrench{}{}:=\begin{bmatrix}{\bf m}\\{\bf f}\end{bmatrix}\in{\mathbb R}^6$.\\
${\bf K}$ & $6\times6$ stiffness matrix, $\wrench{}{}={\bf K}\disp{}{}$.\\
${\bf u}$ & Coordinates of a point on a surface patch, ${\bf u}\in{\mathbb R}^2$.\\
$\curvature{}$ & $2\times2$ curvature form at a point on a surface.\\
$\torsion{}$ & $1\times2$ torsion form at a point on a surface.\\
$\metric{}$ & $2\times2$ diagonal metric at a point on a surface.\\
$\phi$ & Angle of rotation about $\hat{z}$-axis of frame \{l$_1$\}/\{c$_1$\},\\
& from $\hat{x}$-axis of \{l$_1$\}/\{c$_1$\} to that of \{l$_2$\}/\{c$_2$\}.\\
$m$ & Object mass.\\
$\mathfrak g$ & $3\times1$ gravity vector.\\
\hline
\end{tabular}
\label{table:notation}
\end{table}

\subsection{Preliminaries}
\label{ssec:preliminaries}
\subsubsection{Properties of skew-symmetric matrices}
The $so(3)$ matrix representation $[{\bf a}]$ of ${\bf a} = (a_1,a_2,a_3)$ is 
\begin{equation*}
    [{\bf a}] = \begin{bmatrix}
    0 & -a_3 & a_2 \\
    a_3 & 0 & -a_1 \\
    -a_2 & a_1 & 0
    \end{bmatrix}.
\end{equation*}
These skew-symmetric matrices satisfy
\begin{equation} \label{eq:skew_skew}
[[\bf{a}]\bf{b}]=[\bf{a}][\bf{b}]-[\bf{b}][\bf{a}]~\forall\bf{a},\bf{b}\in\mathbb{R}^3
\end{equation}
and 
\begin{equation} \label{eq:4skews}
\begin{bmatrix}
[\bf{a}] & [\bf{b}]\\
\bf{0}_{3\times3} & [\bf{a}]
\end{bmatrix}
\begin{bmatrix}
\bf{x}\\
\bf{y}
\end{bmatrix}
=-
\begin{bmatrix}
[\bf{x}] & [\bf{y}]\\
[\bf{y}] & \bf{0}_{3\times3}
\end{bmatrix}
\begin{bmatrix}
\bf{a}\\
\bf{b}
\end{bmatrix}
~\forall\bf{a},\bf{b},\bf{x},\bf{y}\in\mathbb{R}^3.
\end{equation} 
A rotation matrix $\R{} \in SO(3)$ and position vector $\p{}{} \in \real^3$ satisfy
\begin{equation} \label{eq:skew_Rp}
[\R{}\p{}{}]=\R{}[\p{}{}]\R{}^\intercal.
\end{equation}

\subsubsection{Transformations}
The representations of the relative position between frames \{i\} and \{j\} satisfy
\begin{equation} \label{eq:p_rep}
\p{i}{ij}=-\p{i}{ji}=-\R{\text{ij}}\p{j}{ji},
\end{equation}
and representations of exponential coordinates, twists, and wrenches satisfy 
\begin{equation} \label{eq:trans_XVF}
\disp{i}{}=\AdT{\text{ij}}\disp{j}{},~
\twist{i}{}=\AdT{\text{ij}}\twist{j}{},~\wrench{i}{}=\AdT{\text{ji}}^\intercal\wrench{j}{}.
\end{equation}
We define a $6 \times 6$ matrix representation $\W{}{}$ of a wrench $\wrench{}{}$ as
\[
\W{}{}:=\begin{bmatrix}{[\bf{m}]}&{[\bf{f}]} \\
{[\bf{f}]}&\bf{0}_{3\times3}\end{bmatrix},
\]
which shares the same units as a stiffness matrix ${\bf K}$. Transformations of ${\bf K}$ and $\W{}{}$ between frames \{i\} and \{j\} satisfy
\begin{equation} \label{eq:trans_KW}
{\bf K}^\text{i}=\AdT{\text{ji}}^\intercal{\bf K}^\text{j}\AdT{\text{ji}},~\W{i}{}=\AdT{\text{ji}}^\intercal\W{j}{}\AdT{\text{ji}}.
\end{equation}

\subsubsection{Time derivatives}
The time derivative of ${\bf T}_\text{ij}$ is
\begin{equation} \label{eq:dT}
\dot{\bf T}_\text{ij}=[\twist{i}{ij}]{\bf T}_\text{ij}~\Rightarrow~\dot{\bf R}_\text{ij}=[\ww{i}{ij}]\R{\text{ij}},~\dot{\bf p}^\text{i}_\text{ij}=[\ww{i}{ij}]\p{i}{ij}+\vv{i}{ij}.
\end{equation}
The time derivative of $\AdT{\text{ij}}$, using~\eqref{eq:skew_skew}, \eqref{eq:skew_Rp}, \eqref{eq:p_rep}, and \eqref{eq:dT}, is 
\begin{equation} \label{eq:dt_adj}
\frac{d}{dt}(\AdT{\text{ij}})=-\AdT{\text{ij}}
\begin{bmatrix}
[\ww{j}{ji}] & \bf{0}_{3\times3}\\
[\vv{j}{ji}] & [\ww{j}{ji}]
\end{bmatrix}.
\end{equation}
Details are in Appendix \ref{cha:derivation_dt_adj}.

For each finger, the flexure stiffness represented in the frame \{f$_0$\}, $\stiffness{\text{f}_0}$, is a constant matrix, and the wrench applied to the spring is
${\mathcal F}^{\text{f}_0}_\text{spr}= \stiffness{\text{f}_0}\disp{f$_0$}{f$_0$f}$ for small displacements $\disp{f$_0$}{f$_0$f}$. The time derivative is
\begin{equation*}
\dot{\mathcal F}^{\text{f}_0}_\text{spr}=\frac{d}{dt}(\stiffness{\text{f}_0}\disp{f$_0$}{f$_0$f})=\stiffness{\text{f}_0}\dot{\mathcal X}^{\text{f}_0}_{\text{f}_0\text{f}}.
\end{equation*}
For small displacements, $\dot{\mathcal X}^{\text{f}_0}_{\text{f}_0\text{f}}\approx\twist{f$_0$}{f$_0$f}$. Substituting this yields
\begin{equation} \label{eq:dt_F}
\dot{\mathcal F}^{\text{f}_0}_\text{spr}=\stiffness{\text{f}_0}\twist{f$_0$}{f$_0$f}.
\end{equation}
The time derivative of a wrench $\dot{\mathcal{F}}^{\text{i}}$ can be represented in another moving frame \{j\} using~\eqref{eq:4skews} and \eqref{eq:dt_adj} (details in Appendix \ref{cha:derivation_dt_wrench}): 
\begin{equation} \label{eq:dt_wrench}
\dot{\mathcal{F}}^{\text{i}}=\frac{d}{dt}(\AdT{\text{ji}}^\intercal\wrench{j}{})=-\W{i}{}\twist{i}{ij}+\AdT{\text{ji}}^\intercal\dot{\mathcal{F}}^{\text{j}}.
\end{equation}

\subsubsection{Contact kinematics}
When a fingertip rolls, spins, and/or slides on an object's surface while maintaining contact with the object, the twist representing the object's motion relative to the fingertip can be represented as
\begin{equation*}
\twist{l$_1$}{l$_1$l$_2$}=\begin{bmatrix}
\omega_x\\\omega_y\\\omega_z\\v_x\\v_y\\v_z
\end{bmatrix}
\begin{tabular}{cl}
\multirow{2}{*}{$\}\rightarrow$} & \multirow{2}{*}{pure rolling}\\
&\\
$\rightarrow$ & spinning about $\hat{z}$-axis\\
\multirow{2}{*}{$\}\rightarrow$} & \multirow{2}{*}{sliding}\\
&\\
$\equiv 0$ & impenetrability
\end{tabular}
\label{eq:states}
\end{equation*}

Following the contact kinematics derivation from Montana~\cite{montana1988kinematics},
\begin{equation} \label{eq:montana}
\begin{split}
&\vv{c$_1$}{oc$_1$}=
\begin{bmatrix}\metric{,1}\dot{\bf u}_1\\0\end{bmatrix},\\
&[\ww{c$_1$}{oc$_1$}]=
\begin{bmatrix}
0 & -\torsion{,1}\metric{,1}\dot{\bf u}_1 & \multirow{2}{*}{$\curvature{,1}\metric{,1}\dot{\bf u}_1$}\\
\torsion{,1}\metric{,1}\dot{\bf u}_1 & 0 &\\
\multicolumn{2}{c}{-(\curvature{,1}\metric{,1}\dot{\bf u}_1)^\intercal} & 0
\end{bmatrix}\\
&\vv{c$_2$}{fc$_2$}=
\begin{bmatrix}\metric{,2}\dot{\bf u}_2\\0\end{bmatrix},\\
&[\ww{c$_2$}{fc$_2$}]=
\begin{bmatrix}
0 & -\torsion{,2}\metric{,2}\dot{\bf u}_2 & \multirow{2}{*}{$\curvature{,2}\metric{,2}\dot{\bf u}_2$}\\
\torsion{,2}\metric{,2}\dot{\bf u}_2 & 0 &\\
\multicolumn{2}{c}{-(\curvature{,2}\metric{,2}\dot{\bf u}_2)^\intercal} & 0
\end{bmatrix}\\
&\vv{c$_2$}{c$_2$c$_1$}={\bf 0}_{3\times1},~
[\ww{c$_2$}{c$_2$c$_1$}]=
\begin{bmatrix}
0 & -\dot{\phi} & 0\\
\dot{\phi} & 0 & 0\\
0 & 0 & 0
\end{bmatrix}.
\end{split}   
\end{equation}
Defining the $6\times2$ matrix
\begin{equation*}
{\bf G}_i:=
\begin{bmatrix}
\begin{bmatrix}0&-1\\1&0\end{bmatrix}
\curvature{,i}\\
\torsion{,i}\\
{\bf I}_{2\times2}\\
{\bf 0}_{1\times2}
\end{bmatrix},~i=1,2,
\end{equation*}
Montana's Equation~\eqref{eq:montana} can be rewritten as
\begin{equation} \label{eq:V_montana}
\begin{cases}
\twist{c$_1$}{oc$_1$}={\bf G}_1\metric{,1}\dot{\bf u}_1,\\
\twist{c$_2$}{fc$_2$}={\bf G}_2\metric{,2}\dot{\bf u}_2,\\
\twist{c$_2$}{c$_2$c$_1$}=[0,0,\dot{\phi},0,0,0]^\intercal.
\end{cases}
\end{equation}
Moreover, $\dot{\bf u}_1$ can be expressed by elements of $\twist{l$_1$}{l$_1$l$_2$}$ as
\begin{equation}
\dot{\bf u}_1=\metric{,1}^{-1}(\curvature{,1}+\widetilde{\bf K}_\text{cur.,2})^{-1}(
\begin{bmatrix}\omega_y\\-\omega_x\end{bmatrix}
+\widetilde{\bf K}_\text{cur.,2}
\begin{bmatrix}v_x\\v_y\end{bmatrix}
),
\label{eq:u1}
\end{equation}
where 
\begin{equation*}
\widetilde{\bf K}_\text{cur.,2}:=\R{\phi}\curvature{,2}\R{\phi},~\R{\phi}:=
\begin{bmatrix}
\cos\phi & \sin\phi\\
\sin\phi & -\cos\phi
\end{bmatrix}.
\end{equation*}
To maintain contact, $v_z=0$. In the absence of sliding, $v_x=v_y=0$. Substituting these into \eqref{eq:u1} yields
\begin{equation*}
\metric{,1}\dot{\bf u}_1=(\curvature{,1}+\widetilde{\bf K}_\text{cur,2})^{-1}{\bf N}\twist{l$_1$}{l$_1$l$_2$},~{\bf N}:=
\begin{bmatrix}
0 & 1 & \multirow{2}{*}{${\bf 0}_{2\times4}$}\\
-1 & 0 &
\end{bmatrix}.
\end{equation*}
Further substituting into \eqref{eq:V_montana} yields the twist of the object contact frame relative to the object frame,
\begin{equation} \label{eq:Voc1_curv}
\twist{c$_1$}{oc$_1$}={\bf L}_1\twist{l$_1$}{l$_1$l$_2$},~{\bf L}_1:={\bf G}_1(\curvature{,1}+\widetilde{\bf K}_\text{cur,2})^{-1}{\bf N}.
\end{equation}
Equation~\eqref{eq:Voc1_curv} embeds the contact kinematics derivation from Montana for rolling contact, and it shows the relation between $\twist{c$_1$}{oc$_1$}$ and $\twist{l$_1$}{l$_1$l$_2$}$, which will be used in the next section.

\section{Contact Mechanics and Control of In-hand Rolling Manipulation} 
\label{sec:multi_fingers}
\noindent A rigid object of known mass and geometry is grasped by $n$ compliant fingers. The object is in point contact with each fingertip and can roll or spin on the fingertip's surface without sliding. An external wrench $\wrench{}{\text{ext}}$ acts on the object due to gravity. If the object were in contact with an environment, wrenches due to those contacts would also appear in $\wrench{}{\text{ext}}$. For simplicity, we focus on the case of in-hand manipulation of an object in free space.

The controls for the quasistatic hand-object system are the twists of the finger anchors. The state is defined as the configuration of the object relative to a world frame fixed to the palm, the contact points on the object, the corresponding contact frames at the contact points on the fingertips (which, together with the object configuration and contact points, specify the configurations of the fingertips), and the finger anchor configurations. This description uniquely defines the configuration of each fingertip relative to its anchor, and the stiffness of the finger flexure yields the wrench
\(\wrench{}{con,$i$}\) applied to the object by fingertip $i$. For a state to be a valid quasistatic grasp, 
\begin{equation} \label{eq:wrench_equilib}
\sum_{i=1}^{n}\wrench{}{con,$i$}+\wrench{}{ext}={\bf 0}_{6\times1}.
\end{equation}

In this section, we derive the quasistatic forward mechanics (given the state and controls, determine the rate of change of the state) and a version of the inverse mechanics (given the state and the desired object twist, determine consistent controls). 
Using the inverse mechanics, we propose an in-hand rolling manipulation controller based on vision and tactile feedback. 

\subsection{Mechanics constraints}

The forward and inverse rolling mechanics must satisfy no-slip and wrench equilibrium constraints. Below, we derive these constraints and collect them together into a set of equations that forms the basis for the forward and inverse mechanics.

\subsubsection{Rolling constraints}
For each contact, the rolling constraint indicates there is no relative linear velocity between the object and the fingertip
\begin{equation*}
\begin{split}
{\bf v}^{\text{l}_1}_{\text{l}_1\text{l}_2}=
\begin{bmatrix}
\bf{0}_{3\times3} & \bf{I}_{3\times3}    
\end{bmatrix}
\twist{l$_1$}{l$_1$l$_2$}
&= \\
\begin{bmatrix}
\bf{0}_{3\times3} & \bf{I}_{3\times3}    
\end{bmatrix}
\cancelto{\bf I}{\AdT{{\text l}_1{\text c}_1}}\quad\AdT{{\text c}_1{\text w}}\twist{}{l$_1$l$_2$}&=\\
\begin{bmatrix}
[\p{c$_1$}{c$_1$w}]\R{\text{c}_1\text{w}} & \R{\text{c}_1\text{w}} 
\end{bmatrix}
(\cancelto{\bf 0}{\twist{}{l$_1$o}}+\twist{}{ow}+\twist{}{wf}+\cancelto{\bf 0}{\twist{}{fl$_2$}}~)
&=\bf{0}_{3\times1}.
\end{split}
\end{equation*}
Using~\eqref{eq:skew_Rp} and \eqref{eq:p_rep} to rewrite yields
\begin{equation} \label{eq:rolling_final}
{\bf P}_i(\twist{}{wf,$i$}-\twist{}{wo})={\bf 0}_{3\times1},~\forall i=1,\dots,n,
\end{equation}
where ${\bf P}_i:=\begin{bmatrix}
[\p{}{wc$_{1,i}$}] & -\bf{I}_{3\times3}
\end{bmatrix} \in \real^{3\times6}$ and ${\bf p}_{\text{wc}_{1,i}}={\bf p}_{\text{wc}_{2,i}}$ is the location of each fingertip contact.

\subsubsection{Wrench equilibrium}\label{subsec:force_equ}
The contact wrench $\wrench{}{con}$ applied by a fingertip to an object is equal to the negative of the wrench applied to the fingertip flexure,
\begin{equation} \label{eq:f_balance}
\wrench{}{con} = -\wrench{}{spr} ~\Leftrightarrow~ {\bf W}^{}_\text{con} = - {\bf W}^{}_\text{spr}.
\end{equation}
Taking the time derivative of $\wrench{c$_1$}{con}$ and substituting~\eqref{eq:dt_F}, \eqref{eq:dt_wrench}, and \eqref{eq:f_balance} yields
\begin{equation} \label{eq:dt_f_con}
\begin{split}
\dot{\mathcal F}^{\text{c}_1}_\text{con}= &-\dot{\mathcal F}^{\text{c}_1}_\text{spr}\\
=&-{\bf W}^{\text{c}_1}_\text{con}(\twist{c$_1$}{c$_1$o}+\twist{c$_1$}{ow}+\twist{c$_1$}{wa}+\cancelto{\bf 0}{\twist{c$_1$}{af$_0$}}~)-\AdT{\text{f}_0\text{c}_1}^\intercal\dot{\mathcal F}^{\text{f}_0}_\text{spr}\\
=&{\bf W}^{\text{c}_1}_\text{con}(\twist{c$_1$}{oc$_1$}+\twist{c$_1$}{wo}-\twist{c$_1$}{wa})-\AdT{\text{f}_0\text{c}_1}^\intercal\stiffness{\text{f}_0}\twist{f$_0$}{f$_0$f},
\end{split}
\end{equation}
where the fingertip displacement rate $\twist{f$_0$}{f$_0$f}$ satisfies
\begin{equation*}
\twist{f$_0$}{f$_0$f}=\cancelto{\bf 0}{\twist{f$_0$}{f$_0$a}}+\twist{f$_0$}{aw}+\twist{f$_0$}{wf}.
\end{equation*}
Further substituting this into~\eqref{eq:dt_f_con} and transforming using~\eqref{eq:trans_XVF} and \eqref{eq:trans_KW} yields
\begin{equation} \label{eq:df_con}
\begin{split}
\dot{\mathcal{F}}^{\text{c}_1}_\text{con}=&\AdT{\text{w}\text{c}_1}^\intercal{\bf W}_\text{con}(\twist{}{oc$_1$}+\twist{}{wo}-\twist{}{wa})\\
&-\AdT{\text{w}\text{c}_1}^\intercal\stiffness{}(\twist{}{aw}+\twist{}{wf}))\\
=&\AdT{\text{w}\text{c}_1}^\intercal((\stiffness{}-{\bf W}_\text{con})\twist{}{wa}-\stiffness{}\twist{}{wf}\\
&+{\bf W}_\text{con}(\twist{}{wo}+\twist{}{oc$_1$})),
\end{split}
\end{equation}
where $\twist{}{oc$_1$}$ can be expressed using~\eqref{eq:Voc1_curv} as
\begin{equation} \label{eq:V_oc1}
\begin{split}
\twist{}{oc$_1$}&=\AdT{\text{wc}_1}{\bf L}_1\twist{l$_1$}{l$_1$l$_2$}=\AdT{\text{wc}_1}{\bf L}_1\AdT{\text{c}_1\text{w}}(\twist{}{ow}+\twist{}{wf})\\
&=\widetilde{\bf L}_1(\twist{}{wf}-\twist{}{wo})
\end{split}
\end{equation}
and $\widetilde{\bf L}_1:=\AdT{\text{wc}_1}{\bf L}_1\AdT{\text{wc}_1}^{-1} \in \real^{6\times6}$, in which ${\bf R}_{\text{wc}_1}$ depends on the local surface patch and coordinates at the contact on the object. Substituting~\eqref{eq:V_oc1} into~\eqref{eq:df_con} yields
\begin{equation} \label{eq:f_con_mf}
\begin{split}
\dot{\mathcal{F}}^{\text{c}_1}_{\text{con},i}=&[\dot{\bf m}^{\text{c}_1}_{\text{con},i},\dot{\bf f}^{\text{c}_1}_{\text{con},i}]^\intercal\\
=&\AdT{\text{w}\text{c}_{1,i}}^\intercal(-{\bf A}_i\twist{}{wf,$i$}-{\bf B_i}\twist{}{wo}+{\bf C}_i\twist{}{wa,$i$})\\
&\forall i=1, \dots, n,
\end{split}
\end{equation}
where
\begin{equation*}
\begin{cases}
{\bf A}_i:={\bf K}_{\text{spr},i}-{\bf W}_{\text{con},i}\widetilde{\bf L}_{1,i} \in \real^{6\times6}\\
{\bf B}_i:={\bf W}_\text{con,$i$}(\widetilde{\bf L}_{1,i}-{\bf I}_{6\times6}) \in \real^{6\times6}\\
{\bf C}_i:={\bf K}_{\text{spr},i}-{\bf W}_\text{con,$i$} \in \real^{6\times6}.
\end{cases}
\end{equation*}

For point contacts with no torsional friction, the moment portion of the contact wrench is always zero, so $\dot{\bf m}^{\text{c}_1}_\text{con}\equiv{\bf 0}_{3\times1}$, which indicates
\begin{equation} \label{eq:fingertips_final}
{\bf Q}_i({\bf A}_i\twist{}{wf,$i$}+{\bf B}_i\twist{}{wo}-{\bf C}_i\twist{}{wa,$i$})={\bf 0}_{3\times1}, ~\forall i=1,\dots,n,
\end{equation} 
where ${\bf Q}_i:=\begin{bmatrix}
-{\bf I}_{3\times3} & [\p{}{wc$_{1,i}$}]
\end{bmatrix}\in \real^{3\times6}$. 

Finally, taking the derivative of the wrench balance constraint \eqref{eq:wrench_equilib} and substituting~\eqref{eq:dt_wrench} and \eqref{eq:df_con} yields 
\begin{equation} \label{eq:object_final}
\sum_{i=1}^{n}({\bf K}_{\text{spr},i}\twist{}{wf,$i$})=\sum_{i=1}^{n}({\bf C}_i\twist{}{wa,$i$})+\dot{\mathcal F}_\text{ext}.
\end{equation}
Details are in Appendix \ref{cha:derivation_object_final}. When there is no environmental contact, ${\mathcal F}_\text{ext}$ is determined by the object pose, mass, and gravity,
\begin{equation*}
{\mathcal F}_\text{ext}=\begin{bmatrix}
{\bf p}_\text{wo}\times m{\mathfrak g}\\
m{\mathfrak g}
\end{bmatrix}=m\begin{bmatrix}
[{\bf p}_\text{wo}]{\mathfrak g}\\
{\mathfrak g}
\end{bmatrix}.
\end{equation*}
Taking the derivative yields
\begin{equation} \label{eq:F_ext}
\begin{split}
\dot{\mathcal F}_\text{ext}&=m\begin{bmatrix}
[\dot{\bf p}_\text{wo}]{\mathfrak g}\\
{\bf 0}_{3\times1}
\end{bmatrix}=m\begin{bmatrix}
-[{\mathfrak g}]([{\boldsymbol\omega}_\text{wo}]{\bf p}_\text{wo}+{\bf v}_\text{wo})\\
{\bf 0}_{3\times1}
\end{bmatrix}\\
&=m\begin{bmatrix}
[{\mathfrak g}]\begin{bmatrix}
[{\bf p}_\text{wo}] & -{\bf I}_{3\times3}
\end{bmatrix}\\
{\bf 0}_{3\times6}
\end{bmatrix}{\mathcal V}_\text{wo}.
\end{split}
\end{equation}
\subsubsection{Combining the constraints}
Stacking~\eqref{eq:rolling_final}, \eqref{eq:fingertips_final}, \eqref{eq:object_final}, and \eqref{eq:F_ext} yields
\begin{equation} \label{eq:forumulation}
{\bf D}_\text{f}{\bf V}_\text{f}+{\bf D}_\text{o}{\cal V}_\text{wo}={\bf D}_\text{a}{\bf V}_\text{a},
\end{equation}
where
\begin{equation*}
\begin{cases}
{\bf D}_\text{f}:=
\begin{bmatrix}
\begin{bmatrix}{\bf Q}_1{\bf A}_1\\{\bf P}_1\end{bmatrix} & \cdots & {\bf 0}_{6\times6} \\
\vdots & \ddots & \vdots \\
{\bf 0}_{6\times6} & \cdots & \begin{bmatrix}{\bf Q}_n{\bf A}_n\\{\bf P}_n\end{bmatrix} \\
{\bf K}_{\text{spr},1} & \cdots & {\bf K}_{\text{spr},n} 
\end{bmatrix} \in \real^{6(n+1)\times 6n}
\\ 

{\bf D}_\text{o}:=
\begin{bmatrix}
\begin{bmatrix}{\bf Q}_1{\bf B}_1\\{\bf -P}_1\end{bmatrix}\\
\vdots\\
\begin{bmatrix}{\bf Q}_n{\bf B}_n\\{\bf -P}_n\end{bmatrix}\\
m[{\mathfrak g}]\begin{bmatrix}
-[{\bf p}_\text{wo}] & {\bf I}_{3\times3}
\end{bmatrix}\\
{\bf 0}_{3\times6}
\end{bmatrix} \in \real^{6(n+1)\times 6}\\

{\bf D}_\text{a}:=
\begin{bmatrix}
\begin{bmatrix}{\bf Q}_1{\bf C}_1\\{\bf 0}_{3\times6}\end{bmatrix} & \cdots & {\bf 0}_{6\times6}\\
\vdots & \ddots & \vdots\\
{\bf 0}_{6\times6} & \cdots & \begin{bmatrix}{\bf Q}_n{\bf C}_n\\{\bf 0}_{3\times6}\end{bmatrix}\\
{\bf C}_1 & \cdots & {\bf C}_n
\end{bmatrix} \in \real^{6(n+1)\times 6n}\\

{\bf V}_\text{f}:=
\begin{bmatrix}
\twist{}{wf,$1$}\\
\vdots\\
\twist{}{wf,$n$}
\end{bmatrix}, {\bf V}_\text{a}:=\begin{bmatrix}\twist{}{wa,$1$}\\\vdots\\\twist{}{wa,$n$}\end{bmatrix} \in \real^{6n},
\end{cases}
\end{equation*}
for all $i=1,\dots,n$. The matrices ${\bf D}_\text{f}$, ${\bf D}_\text{a}$, and ${\bf D}_\text{o}$ are all determined by the system state.

Equation~\eqref{eq:forumulation} collects the $6(n+1)$ state-dependent kinematic and wrench constraints relating the external wrench rate and the twists of the object, finger anchors, and fingertips. 

\subsection{Forward mechanics} \label{sec:multi_finger_mech}
The forward mechanics problem is: Given the current state of the grasp and the controls (the anchor twists ${\bf V}_\text{a}$), find the twists of the object and the fingertips.

Rewriting~\eqref{eq:forumulation} yields
\begin{equation} \label{eq:AXB}
{\bf D}{\bf V}={\boldsymbol \beta},
\end{equation}
where
\begin{equation*}
{\bf D}:=\begin{bmatrix}
{\bf D}_\text{f} & {\bf D}_\text{o}
\end{bmatrix}, {\bf V}:= 
\begin{bmatrix}
{\bf V}_\text{f}\\{\cal V}_\text{wo}
\end{bmatrix}, {\boldsymbol \beta}:={\bf D}_\text{a}{\bf V}_\text{a}.
\end{equation*}
There are $6(n+1)$ unknowns, ${\bf V}$, to be solved using the same number of constraints. The rank of ${\bf D}$ depends on the current state; generically, it is full rank, with $\operatorname{rank}({\bf D})=6(n+1)$. In this case, ${\bf D}$ is invertible and~\eqref{eq:AXB} gives a unique solution for the fingertip and object twists, ${\bf V}={\bf D}^{-1}{\boldsymbol \beta}$. 
In singular states where ${\bf D}$ is not full rank, multiple solutions exist. A higher-order analysis including dynamics would be needed to resolve such cases.\footnote{For example, consider a ball squeezed between two flat, parallel, opposing fingertips attached to stationary anchors. Because there is no torsional friction at the contacts, no contact moments can be generated about the common contact normal axis, and there is no force-closure grasp. Any spin of the ball about the common contact normal axis satisfies quasistatic wrench balance and rolling constraints.}

\subsection{Inverse mechanics}
\noindent
The inverse mechanics problem is: Given the current state of the grasp and the desired object twist, find consistent controls (the finger anchor twists ${\bf V}_\text{a}$). 

Based on~\eqref{eq:AXB}, define a selection matrix 
\begin{equation} \label{eq:selection}
\begin{split}
{\bf S}^j_i:=&\begin{bmatrix}
{\bf S}'_1 & \cdots & {\bf S}'_i    
\end{bmatrix} \in \real^{6\times 6i}, \text{where}\\
{\bf S}'_k=&\begin{cases}
{\bf I}_{6\times6},~k=j\\
{\bf 0}_{6\times6},~\text{otherwise}
\end{cases}, 1\leq k\leq i.
\end{split}
\end{equation}
When ${\bf D}$ is invertible, the object twist can be expressed as
\begin{equation} \label{eq:V_a}
\twist{}{wo}={\bf S}^{n+1}_{n+1}{\bf V}={\bf S}^{n+1}_{n+1}{\bf D}^{-1}{\boldsymbol \beta}={\bf \Pi}{\bf V}_\text{a},
\end{equation}
where ${\bf \Pi}:={\bf S}^{n+1}_{n+1}{\bf D}^{-1}{\bf D}_\text{a} \in \real^{6 \times 6(n+1)}$. If ${\bf \Pi}$ is full rank ($\operatorname{rank}({\bf \Pi}) = 6$), then there exists a set of anchor twists consistent with the desired $\twist{}{wo}$. For practical robot hand control, the fingers must be capable of achieving an element of this set. 

\subsection{Contact force inequalities} \label{sec:force_ineq}
The derivation above does not consider contact force inequalities that must be satisfied during rolling manipulation: (1) the normal force must be nonnegative at each fingertip contact and (2) each contact force must satisfy friction limits. Beginning from an initial grasp that satisfies these constraints, we formulate the continued enforcement of these inequality constraints as (1) a constraint on the rate of change of the contact force magnitude when it reaches a minimum value and (2) a constraint on the rate of change of the contact force direction when the force is at the friction limit.

Assuming ${\bf D}$ is invertible, Equations~\eqref{eq:f_con_mf}, \eqref{eq:AXB}, \eqref{eq:selection}, and \eqref{eq:V_a} indicate that the rate of change of the contact force at contact $i$ is 
\begin{equation*} \label{eq:contact_force}
\begin{split}
\dot{\bf f}^{\text{c}_1}_{\text{con},i}=&\begin{bmatrix}
{\bf 0}_{3\times3} & {\bf R}_{{\text wc}_{1,i}}^\intercal
\end{bmatrix}(-{\bf A}_i\twist{}{wf,$i$}-{\bf B_i}\twist{}{wo}
+{\bf C}_i\twist{}{wa,$i$})\\
=&\begin{bmatrix}
{\bf 0}_{3\times3} & {\bf R}_{{\text wc}_{1,i}}^\intercal
\end{bmatrix}\\
&(-{\bf A}_i{\bf S}^i_{n+1}{\bf V}-{\bf B_i}{\bf S}^{n+1}_{n+1}{\bf V}
+{\bf C}_i{\bf S}^i_n{\bf V}_\text{a})\\
=&{\bf \Psi}_{\text{a},i}{\bf V}_\text{a},
\end{split}
\end{equation*}
where
\begin{equation*}
\begin{split}
{\bf \Psi}_{\text{a},i}:=&\begin{bmatrix}
{\bf 0}_{3\times3} & {\bf R}_{{\text wc}_{1,i}}^\intercal
\end{bmatrix}\\
&(-{\bf A}_i{\bf S}^i_{n+1}{\bf D}^{-1}{\bf D}_\text{a}-{\bf B}_i{\bf \Pi}+{\bf C}_i{\bf S}^i_n).
\end{split}
\end{equation*}

When $||{\bf f}^{\text{c}_1}_{\text{con},i}||\leq f_\text{min}$, where $f_\text{min}>0$ is a defined minimum allowed contact force magnitude, we can enforce the constraint that the contact force magnitude cannot decrease, 
${\bf f}^{\text{c}_1\intercal}_{\text{con},i}\dot{\bf f}^{\text{c}_1}_{\text{con},i} \geq0$, i.e.,
\begin{equation} \label{eq:contact_normal}
\end{equation}

To enforce the friction constraint, we define $\mu_{\text{max}}$ as the maximum friction coefficient that the in-hand rolling controller can assume to be available at each fingertip contact ($\mu_{\text{max}}<\mu$, where $\mu$ is the actual friction coefficient). Finger $i$'s contact force is written ${\bf f}^{\text{c}_1}_{\text{con},i} = \begin{bmatrix}
f^{\text{c}_1}_{\text{con},i,x},f^{\text{c}_1}_{\text{con},i,y},f^{\text{c}_1}_{\text{con},i,z}
\end{bmatrix}^\intercal$. Then the friction constraint becomes active if 
\begin{equation} \label{eq:fri_coeff}
\frac{\sqrt{(f^{\text{c}_1}_{\text{con},i,x})^2 + (f^{\text{c}_1}_{\text{con},i,y})^2}}{f^{\text{c}_1}_{\text{con},i,z}}\geq\mu_\text{max},
\end{equation}
and it can be written as
\begin{equation}\label{eq:contact_cone}
\begin{split}
(({\bf H} {\bf f}^{\text{c}_1}_{\text{con},i} \times {\bf f}^{\text{c}_1}_{\text{con},i}) \times {\bf f}^{\text{c}_1}_{\text{con},i})^\intercal\dot{\bf f}^{\text{c}_1}_{\text{con},i}&\geq0, \text{i.e.,}\\
{\bf Z}_\text{con}{\bf \Psi}_{\text{a},i}{\bf V}_\text{a}&\leq0,
\end{split}
\end{equation}
where
\begin{equation*}
\begin{split}
{\bf Z}_\text{con}:=&{\bf f}^{\text{c}_1 \intercal}_{\text{con},i}[[{\bf H} {\bf f}^{\text{c}_1}_{\text{con},i}]{\bf f}^{\text{c}_1}_{\text{con},i}]\\
{\bf H}:=&\begin{bmatrix}
1 & 0 & 0\\
0 & 1 & 0\\
0 & 0 & 0
\end{bmatrix}.
\end{split}
\end{equation*}

\subsection{Calculating finger joint velocities for inverse mechanics}
Of anchor twists ${\bf V}_\text{a}$ satisfying~\eqref{eq:V_a} and contact force inequality constraints, we are specifically interested in anchor twists that can be achieved by the fingers of a given robot hand. Each finger's Jacobian matrix $\bf J$ maps its joint velocities $\dot{\boldsymbol \theta}$ to its anchor twist $\twist{}{wa}={\bf J}\dot{\boldsymbol \theta}$. Stacking for all $n$ fingers yields 
\begin{equation} \label{eq:reachable}
{\bf V}_\text{a}={\bf \Xi}\dot{\bf \Theta},~{\bf \Xi}:=
\begin{bmatrix}
{\bf J}_1 & & {\bf 0}\\
 & \ddots & \\
 {\bf 0} & & {\bf J}_n 
\end{bmatrix}
,~\dot{\bf \Theta}:=
\begin{bmatrix}\dot{\boldsymbol \theta}_1\\\vdots\\\dot{\boldsymbol \theta}_n\end{bmatrix}.
\end{equation}
Substituting into~\eqref{eq:V_a} yields
\begin{equation*} \label{eq:AXB_control}
{\bf \Sigma}\dot{\bf \Theta}=\twist{}{wo},
\end{equation*}
where ${\bf \Sigma} = {\bf \Pi}{\bf \Xi}$ is a $6\times \operatorname{dim}(\dot{\bf \Theta})$ matrix. 
If $\operatorname{rank}({\bf \Sigma}) < 6$, the hand cannot achieve most object twists via in-hand rolling.

When $\operatorname{rank}({\bf \Sigma}) = 6$, we can solve for joint velocities $\dot{\bf \Theta}$ that achieve the desired object twist and minimize its two-norm using the following convex quadratic program:
\begin{equation} \label{eq:QP}
\begin{split}
&\underset{\dot{\bf \Theta}}{\operatorname{min}} ~ \dot{\bf \Theta}^\intercal\dot{\bf \Theta}  \\
\text{s.t.} ~& {\bf \Sigma}\dot{\bf \Theta} =\twist{}{wo}\\
& -{\bf f}^{\text{c}_1\intercal}_{\text{con},i}{\bf \Psi}_{\text{a},i}{\bf V}_\text{a} \leq 0 ~\text{for all~} ||{\bf f}^{\text{c}_1}_{\text{con},i}||<f_{\text{min}}\\
& {\bf Z}_\text{con}{\bf \Psi}_{\text{a},i}{\bf V}_\text{a} \leq0 ~\text{for all~} \eqref{eq:fri_coeff}.
\end{split}
\end{equation}
When no inequality constraints are active, the closed-form solution is
\begin{equation} \label{eq:sol_multi_hand}
\dot{\bf \Theta}= 
{\bf \Sigma}^\dagger\twist{}{wo}, ~\text{where}~{\bf \Sigma}^\dagger = {\bf \Sigma}^\intercal({\bf \Sigma}{\bf \Sigma}^\intercal)^{-1}.
\end{equation}

\subsection{In-hand rolling manipulation controller}
\begin{figure}
    \centering
\includegraphics{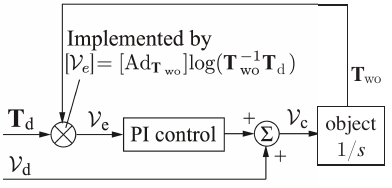}
    \caption{PI-plus-feedforward object trajectory control.}
    \label{fig:obj-control}
\end{figure}

As an application of the inverse mechanics derived above, we can construct an in-hand rolling manipulation trajectory-tracking controller based on the PI-plus-feedforward task-space motion control template of Figure~\ref{fig:obj-control}~\cite{lynch2017modern}. In this template, the desired object pose ${\bf T}_{\text{d}}$ is compared to the actual pose of the object ${\bf T}_\text{wo}$ (e.g., from vision feedback) to generate the $se(3)$ object configuration error
\begin{equation*}
    [\twist{}{e}] = [\text{Ad}_{{\bf T}_{\text{wo}}}] \log({\bf T^{-1}_{\text{wo}}}{\bf T}_{\text{d}}).
\end{equation*}
The corresponding error twist $\twist{}{e}$ is fed into a PI controller, and the output is summed with the feedforward desired object twist $\twist{}{d}$ to generate a commanded twist ${\cal V}_\text{c}$ of the object. 

\begin{figure*}
    \centering
\includegraphics[width=6in]{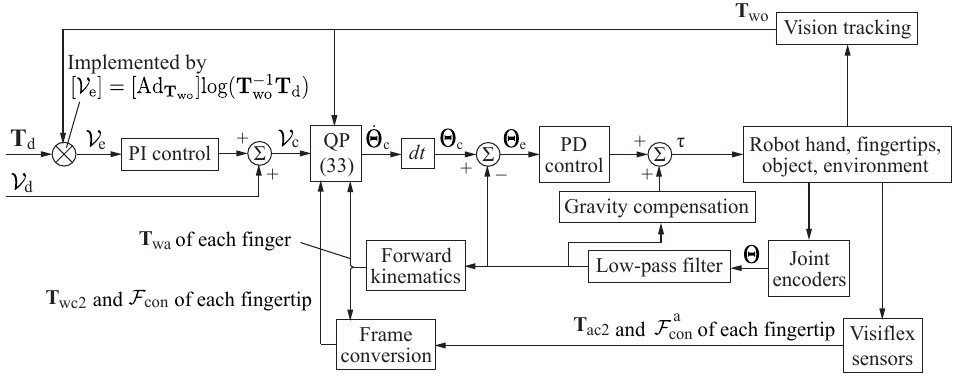}
    \caption{The complete compliant in-hand rolling manipulation feedback controller based on the PI-plus-feedforward control template of Figure~\ref{fig:obj-control}.}
    \label{fig:new-control}
\end{figure*}

Figure~\ref{fig:new-control} illustrates a complete compliant in-hand rolling manipulation controller based on the motion control template of Figure~\ref{fig:obj-control}. The commanded twist ${\cal V}_\text{c}$ is generated by in-hand rolling using the quadratic program~\eqref{eq:QP}, which uses vision feedback, finger encoder feedback, and tactile sensor feedback to calculate finger joint velocities $\dot{\boldsymbol \Theta}_\text{c}$. These joint velocities are numerically integrated to generate commanded finger joint positions ${\boldsymbol \Theta}_\text{c}$, which are tracked using PD control plus gravity compensation (Figure~\ref{fig:new-control}). 

The controller of Figure~\ref{fig:new-control} stabilizes desired object motion using compliant in-hand rolling, but it does not guarantee that a requested object trajectory is feasible. Planning for a feasible trajectory depends on finger joint limits, finger geometry, and other considerations beyond the scope of this paper. 

\section{Simulation and Experiment} \label{cha:experiments} 
\subsection{A simple planar illustration of rolling contact mechanics}
\begin{figure}
\centering
\includegraphics[width=3.4in]{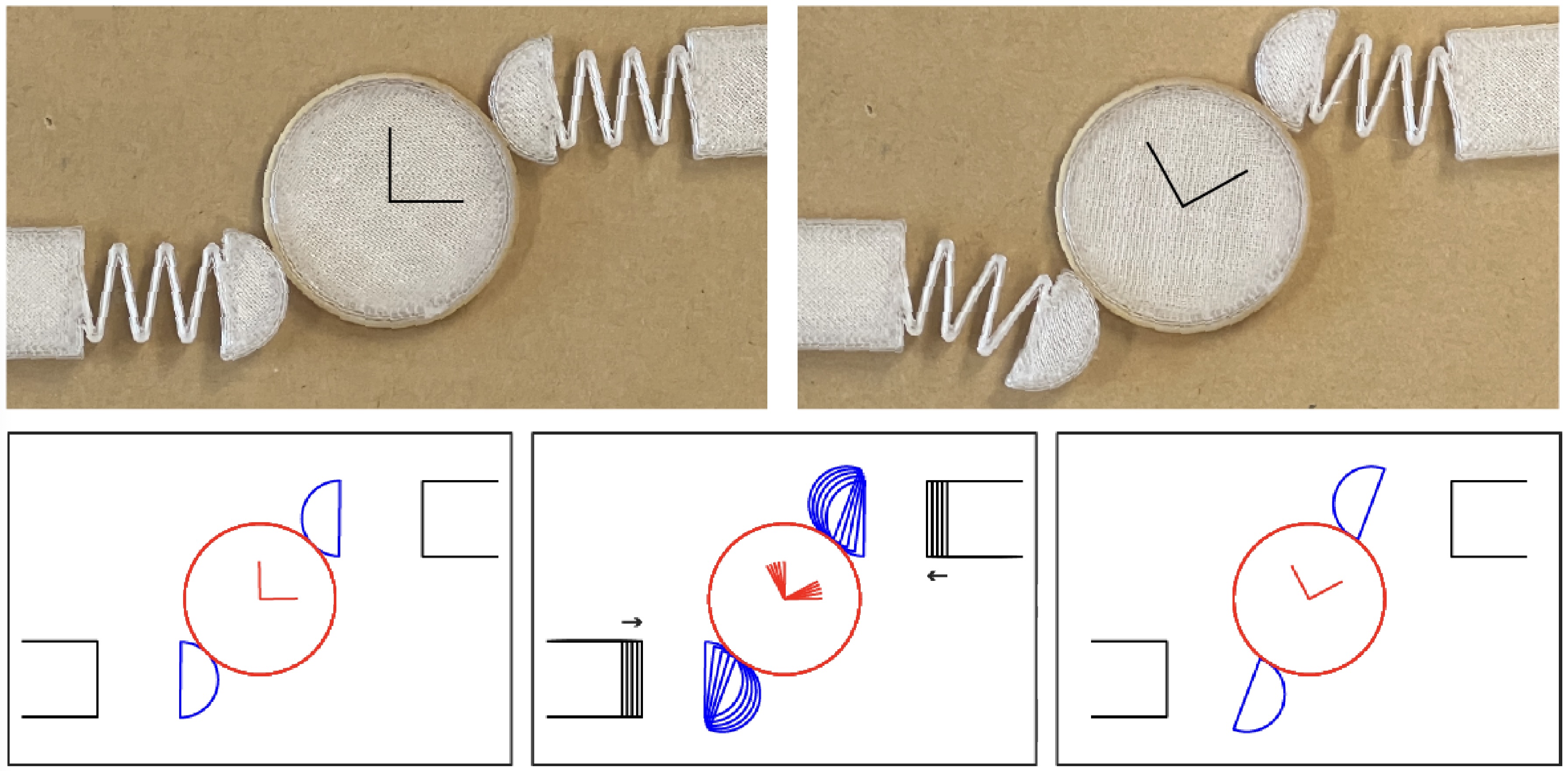}
\caption{(Top left) Two planar fingers in initial contact with a disk. (Top right) When each finger anchor advances forward, the flexure-mounted fingertips comply symmetrically, the fingertips roll on the object, and the disk rotates approximately in place.  (Bottom) A simulation of the manual experiment predicts similar results. Flexures are not rendered in the simulation snapshots.}
\label{fig:sim}
\end{figure}

\noindent 
To provide an intuitive illustration of compliant in-hand rolling, we used a simulation of the forward mechanics~\eqref{eq:AXB} to predict the results of a manual experiment with two 3d-printed planar fingers, each consisting of an anchor, flexure, and fingertip (Figure~\ref{fig:sim}). The fingers manipulate a disk object, and the radii of the circular fingertips and the object are $7.5$~mm and $15$~mm, respectively. A rubber band around the disk increases friction at the contacts.

Figure~\ref{fig:sim} shows the two planar fingers in initial contact with the disk. As each finger anchor advances forward at approximately the same speed, the disk rotates approximately in place, the flexure-mounted fingertips comply symmetrically, and the fingertips roll on the object, as shown in the video. This simple example demonstrates how object and fingertip shapes, flexure properties, and anchor motion work together to generate compliant quasistatic rolling manipulation, as predicted by the forward mechanics~\eqref{eq:AXB}. The simulation of the manual experiment, using estimated stiffnesses of the flexures, yields results similar to the physical experiment, as shown in Figure~\ref{fig:sim}. The figure shows a planar slice of a 3d simulation of a cylindrical object and hemispherical fingertips. 

\begin{table}
\renewcommand{\arraystretch}{1.5}
\centering
\caption{Hardware of the manipulation system}
\begin{tabular}{p{0.7in}p{2.1in}p{0.2in}}
\hline
Component & Inputs and outputs & Rate (Hz)\\
\hline
OptiTrack vision tracking & Output: object pose & 360\\
Tactile sensors & Output: contact locations and wrenches & 60\\
Allegro hand and controller & Input: commanded joint positions. Output: finger configurations and Jacobians & 333\\
WAM arm and controller & Input: all data from other components. Output: finger joint position commands& 500\\
\hline
\end{tabular}
\label{table:hardware}
\end{table}

\subsection{Hardware setup} \label{sec:hardware_setup} % 
\noindent Our experimental setup consists of a Barrett WAM robot arm, an Allegro hand, and four Visiflex tactile fingertips, as shown in Figure~\ref{fig:overview}. We used 10~OptiTrack Prime$^\text{x}$ cameras for object tracking.\footnote{The multi-camera OptiTrack system gives high-resolution, high-frame-rate feedback, but its marker-based method is cumbersome for simple object tracking. Other vision-based tracking methods could easily be subsituted.}
More details are in Table~\ref{table:hardware}. In all experiments, the robot arm and, consequently, the palm were held stationary during manipulation, and all manipulation was performed solely by the fingers.

The Allegro fingers each have four degrees of freedom, meaning that at least two degrees of freedom of relative motion are required at fingertip contacts to achieve general in-hand twists. The rolling contact model in this paper allows up to three degrees of freedom at each contact. At generic grasps away from finger workspace boundaries and contact force inequality constraints, any instantaneous in-hand rolling direction is theoretically achievable by finger anchor motions.

\subsection{Simulation study} \label{sec:sim_study}

\begin{figure}
    \centering
    \includegraphics{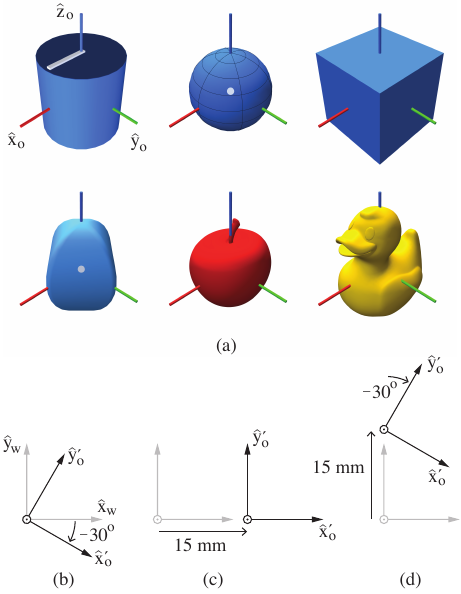}
    \caption{(a) Objects used in the simulation study: cylinder, sphere, cube, ovaloid, apple, and rubber duck. Each object frame semi-axis is $40$~mm long. (b-d) Nominal object trajectories to test tracking control using in-hand rolling manipulation. All trajectories begin with the object frame \{o\} aligned with the world frame \{w\}. In all figures, the $\hat{\text{z}}$-axes of the world and object frames are out of the page. (b)  Rotation about the $\hat{\text{z}}_{\text{w}}$-axis by $-30^\circ$ ($R_z$). (c) Translation along the $\hat{\text{x}}_{\text{w}}$-axis by $15$~mm ($T_x$). (d) Simultaneous translation along the $\hat{\text{y}}_{\text{w}}$-axis by $15$~mm and rotation about the $\hat{\text{z}}_{\text{o}}$-axis by $-30^\circ$ ($T_yR_z$).}
    \label{fig:trajectories}
\end{figure}

To test control based on the compliant rolling mechanics (Figure~\ref{fig:new-control}), we conducted a MuJoCo simulation study of our Visiflex-equipped Allegro hand manipulating six different rigid objects\footnote{The cylinder, sphere, and cube are defined by MuJoCo primitives, and the ovaloid, apple, and rubber duck meshes were adapted from \url{https://github.com/hpfield/Text2Touch}.} along three different desired nominal trajectories starting with the object frame \{o\} coincident with the world frame \{w\} ($\textbf{T}_{\text{wo}}(t = 0) = \textbf{I}$): rotation about the $\hat{\text{z}}_{\text{w}}$-axis by $-30^\circ$ ($R_z$), translation along the $\hat{\text{x}}_{\text{w}}$-axis by $15$~mm ($T_x$), and simultaneous translation along the $\hat{\text{y}}_{\text{w}}$-axis by $15$~mm and rotation about the $\hat{\text{z}}_{\text{o}}$-axis by $-30^\circ$ ($T_yR_z$) (Figure~\ref{fig:trajectories}). All nominal trajectories are at a constant speed with a duration of $5$~s, followed by $5$~s holding the object stationary. The simulation uses a friction coefficient of $\mu = 1.0$ at fingertip contacts and a simulation time step of 0.1~ms.

To generate initial grasps for simulation experiments, the object is initially held stationary with the origin of its frame \{o\} coincident with the world frame \{w\} while three fingers of the hand (thumb, index finger, and middle finger) follow pre-planned joint trajectories to close the fingers on the object. (The fourth finger of the Allegro hand is not used.) When the simulated fingertips contact the fixed object, the fingertip flexures comply, establishing contact forces at the fingertips. After the open-loop finger joint motions complete, the finger joints are held stationary and the object constraints are removed, allowing the object and fingertips to move to reduce the energy stored in the flexures. The object settles into a force-balanced grasp.  This grasp and settling process introduces error in the object's initial configuration relative to the initial configuration of the desired trajectory, and the feedback controller works to eliminate this error.

\subsubsection{Examples}
\begin{figure*}
    \centering
    \includegraphics[width=7in]{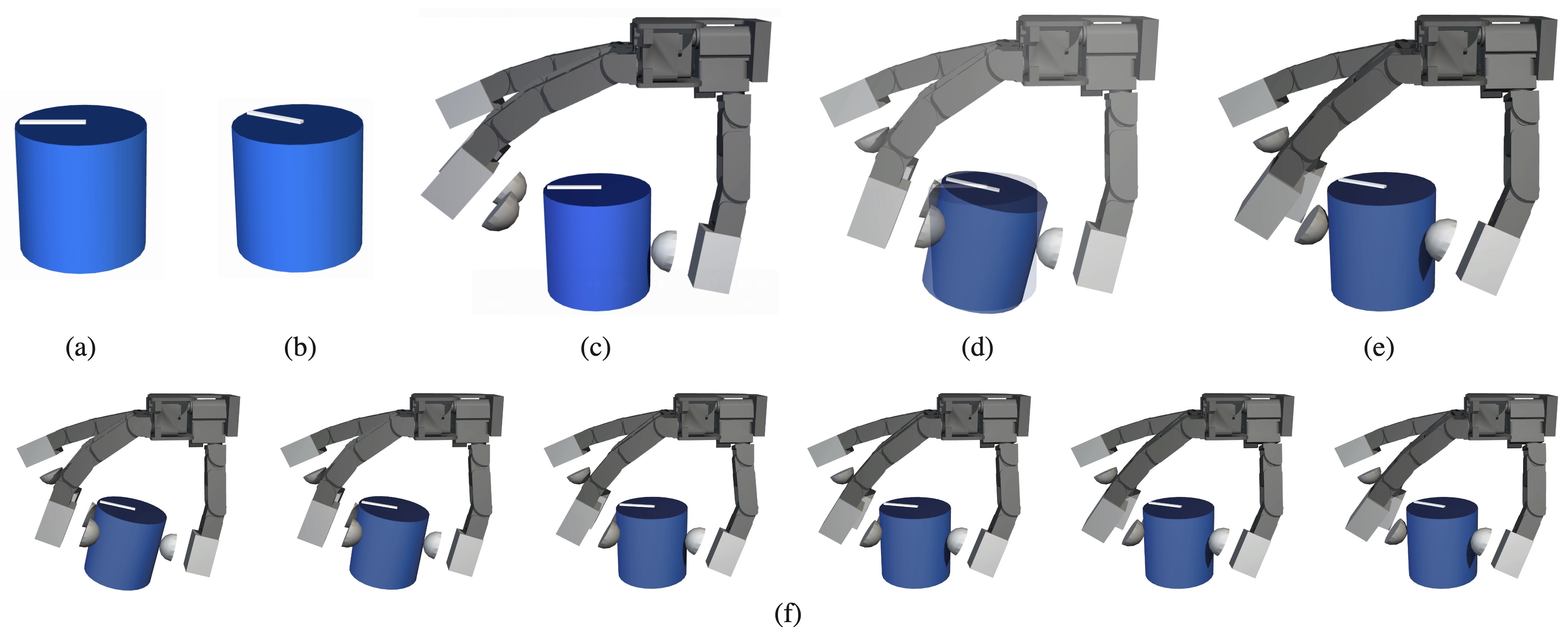}
    \caption{An example of trajectory tracking by compliant in-hand rolling control (Figure~\ref{fig:new-control}). (a) The initial configuration of the cylinder for the planned trajectory. (b) The goal final configuration, $5$~s later, is a rotation by $-30^\circ$ about a vertical axis through the center of the cylinder. (c) Three of the four fingers of the Allegro hand begin to close on the cylinder, which is initially held stationary at the initial configuration of the planned trajectory. The flexures between the finger anchors and the fingertips are not shown. (d) After the fingers finish squeezing and the cylinder settles into a static grasp, the cylinder has moved from the initial configuration of the planned trajectory. This error, primarily a rotation about the $\hat{\text{y}}_{\text{w}}$-axis, must be corrected by the controller. (The starting configuration of the planned trajectory is overlaid to emphasize the error.) (e) After completion of the controlled trajectory, the configuration of the cylinder is visually indistinguishable from the goal configuration. (f) Snapshots at $1$~s intervals during compliant in-hand rolling manipulation, illustrating stabilization of the desired cylinder trajectory over the $5$~s duration of control.}
    \label{fig:basic-control}
\end{figure*}

As a first experiment, we use the cylindrical object and the rotational trajectory $R_z$, as illustrated in Figure~\ref{fig:basic-control} and animated in the video. Figure~\ref{fig:basic-control}(a)-(b) shows the planned initial and final configurations of the cylinder, and Figure~\ref{fig:basic-control}(d) shows the actual initial configuration of the cylinder after a grasp is established by three fingers of the Allegro hand. This initial configuration has approximately $13^\circ$ of orientation error from the planned initial configuration. As illustrated in the snapshots of Figure~\ref{fig:basic-control}(f), however, the cylinder is quickly driven to the desired trajectory by the compliant in-hand rolling controller of Figure~\ref{fig:new-control}. 

\begin{figure}
    \centering
    \includegraphics[width=2.8in]{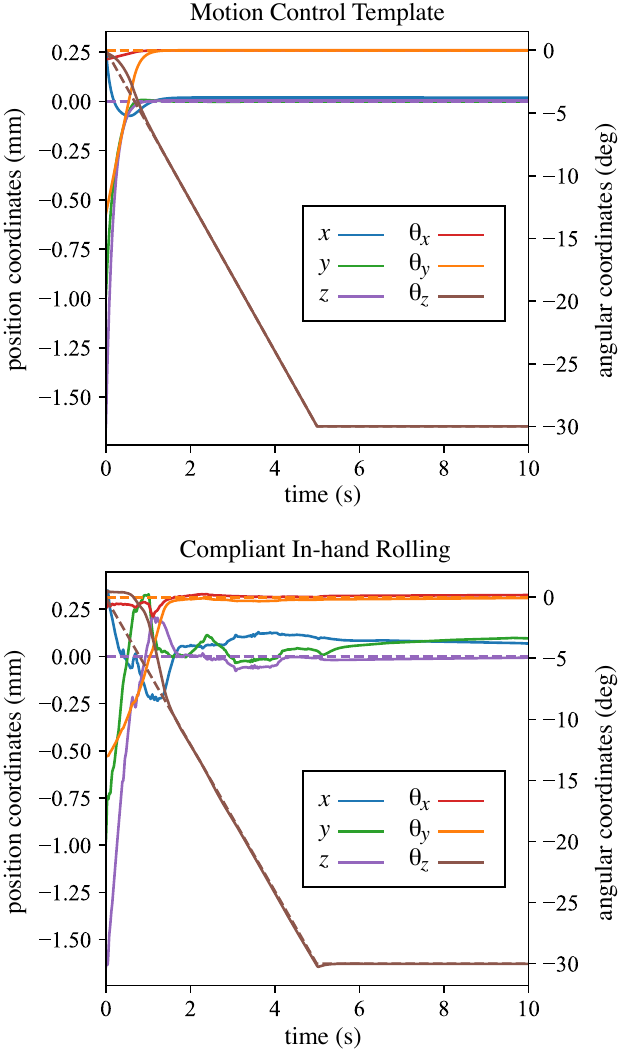}
    \caption{Data corresponding to the task of Figure~\ref{fig:basic-control}. (Top) Direct motion control of the cylinder using the PI-plus-feedforward template of Figure~\ref{fig:obj-control}. This represents the ``ideal'' control that the in-hand rolling controller attempts to emulate when starting from the initial error shown in Figure~\ref{fig:basic-control}. The desired trajectory is zero for all coordinates except $\theta_z$, which ramps from $0$ to $-30^\circ$ over the first $5$~s, then remains constant. (Bottom) The results of simulated in-hand rolling control using the controller of Figure~\ref{fig:new-control}.}
    \label{fig:basic-control-tracking}
\end{figure}

Numerical tracking data are shown in Figure~\ref{fig:basic-control-tracking}.  The top panel shows the ``ideal'' tracking results based on the direct motion control template of Figure~\ref{fig:obj-control}, while the bottom panel shows the results of the simulated manipulation, as illustrated in Figure~\ref{fig:basic-control}. The motion of the cylinder is plotted using the $x$, $y$, and $z$ coordinates of the origin of the cylinder's \{o\} frame relative to \{w\} and the angular exponential coordinates $(\theta_x,\theta_y,\theta_z)$ of \{o\} relative to \{w\}, calculated using $\log \textbf{R}_{\text{wo}}$. The angular coordinate $\theta_z$ quickly converges to the desired ramp from $0$ to $-30^\circ$ while all other coordinates are driven to zero.

\begin{figure}
    \centering
    \includegraphics[width=3.25in]{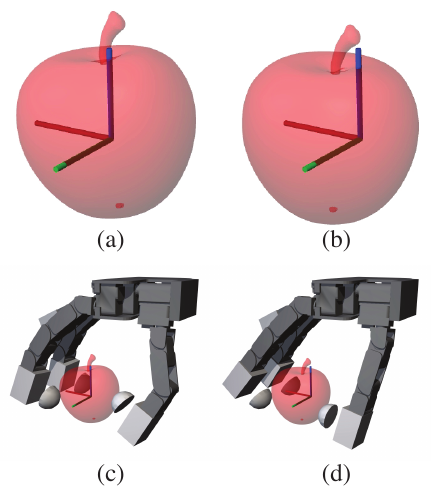}
    \caption{Simulated compliant in-hand rolling of an apple. (a) The initial configuration of the apple. (b) The goal configuration of trajectory $T_x$. (c) The initial configuration after grasping. (d) The configuration after $5$~s of trajectory control, closely achieving the desired apple configuration.}
    \label{fig:apple-motion}
\end{figure}

\begin{figure}
    \centering
    \includegraphics[width=3.25in]{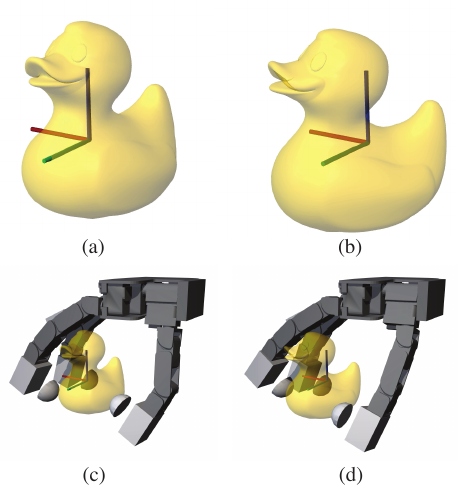}
    \caption{Simulated compliant in-hand rolling of a rubber duck. (a) The initial configuration of the duck. (b) The goal configuration of trajectory $T_yR_z$. (c) The initial configuration after grasping. (d) The configuration after $5$~s of in-hand rolling trajectory control, closely achieving the desired duck configuration.}
    \label{fig:duck-motion}
\end{figure}

Other examples of in-hand rolling manipulation of the apple and the rubber duck, along the $T_x$ and $T_yR_z$ trajectories, are provided in the video and in Figures~\ref{fig:apple-motion} and \ref{fig:duck-motion}, respectively.

In most simulation experiments, across all objects and trajectories, once a stable grasp is achieved, in-hand rolling manipulation is successful in driving the object along the planned trajectory. Exceptions occur when (1) fingers approach their joint limits or self-collide; (2) parts of the fingers other than the fingertips make contact with the object; or (3) the fingertips make contact with concavities (on the duck) or vertices or edges (the cube and the cylinder), where the contact model breaks down. These global considerations are important for initial grasp planning and object trajectory planning, but they are not in the scope of this paper.

\subsubsection{Constrained manipulation}
Although the controller is designed for 6-dof free-space manipulation, the same control scheme can be applied to constrained tasks. In this case, initial grasp acquisition occurs as before, except after the fingers close on the object, the six constraints on the object's motion are reduced to a nonzero number during the object settling phase. The constraints contribute contact forces to the quasistatic wrench balance, and the desired object trajectory must be expressed in the unconstrained space. 

\begin{figure}
    \centering
    \includegraphics[width=2.5in]{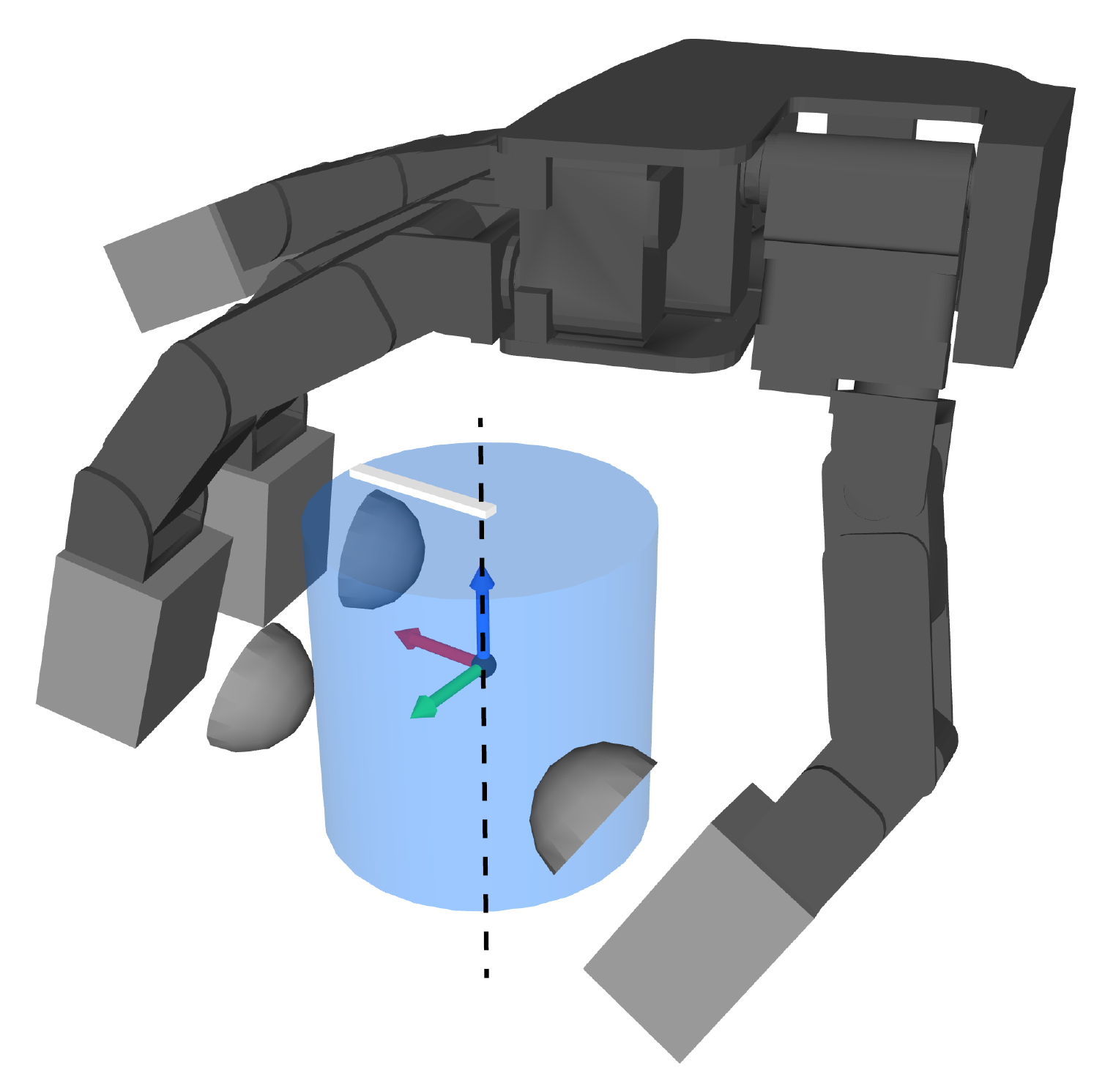}
    \caption{The cylinder constrained to translation and rotation about a vertical axis.}
    \label{fig:constrained-cylinder}
\end{figure}

\begin{figure}
    \centering
    \includegraphics[width=3.2in]{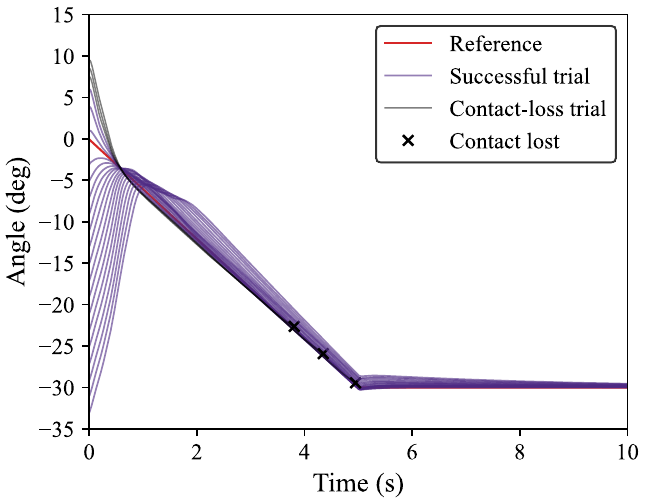}
    \caption{In-hand rolling control of the orientation of a constrained cylinder.}
    \label{fig:constrained-tracking}
\end{figure}

As an example, the cylinder can be constrained to 2-dof rotation about and translation along its $\hat{\text{z}}_{\text{o}}$-axis (Figure~\ref{fig:constrained-cylinder}).  For this example, the cylinder is free to move in $z$ and $\theta_z$, and the desired trajectory is the $R_z$ trajectory, a ramp from $\theta_z = 0^\circ$ to $\theta_z = -30^\circ$. Figure~\ref{fig:constrained-tracking} illustrates the tracking results for initial angles of the cylinder between $-35^\circ$ and $10^\circ$, showing that in-hand rolling control drives convergence to the desired trajectory. In cases where the initial configuration of the cylinder was between $7^\circ$ and $10^\circ$, where the net required cylinder rotation was largest, the fingers lost contact with the cylinder just before the end of the trajectory due to global considerations described above.

In this example, the gain associated with $z$ errors is zero, and the height of the cylinder drifts up or down by up to three millimeters during control.

\subsubsection{Closed-form vs.\ QP-based control}
When the grasp contact forces are away from friction and force magnitude inequality constraints, the closed-form solution~\eqref{eq:sol_multi_hand} is equivalent to the solution provided by the quadratic program~\eqref{eq:QP}. The advantage of the closed-form solution is potentially faster computation times, while the advantage of the QP approach is its ability to handle force inequality constraints.

Testing on a Lenovo 21RQ001HUS with an Intel Core Ultra~7 255HX (20 cores, 20 logical processors) and Windows build 26200 using Python 3.11.9 and NumPy 2.4.4, over thousands of calls of~\eqref{eq:QP} and \eqref{eq:sol_multi_hand} for different combinations of objects and test trajectories, we found that the execution time for the closed-form calculation ranged from $0.5$ to $2.0$~ms, while the execution time for the QP ranged from $0.8$ to $2.8$~ms. 

\begin{figure}
    \centering
\includegraphics[width=1.7in]{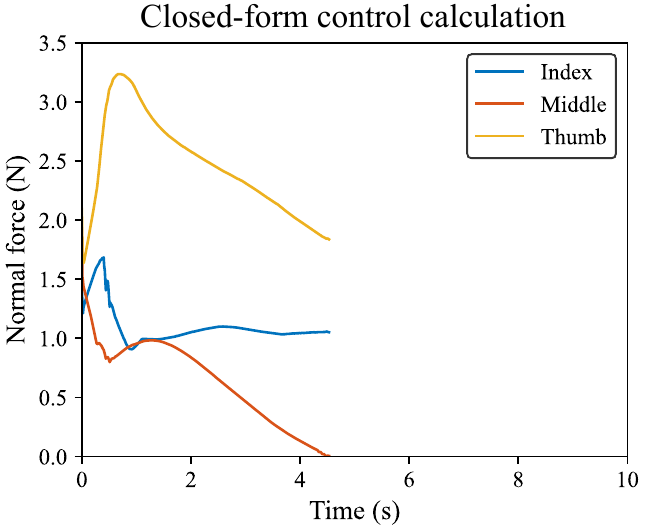} 
\includegraphics[width=1.7in]{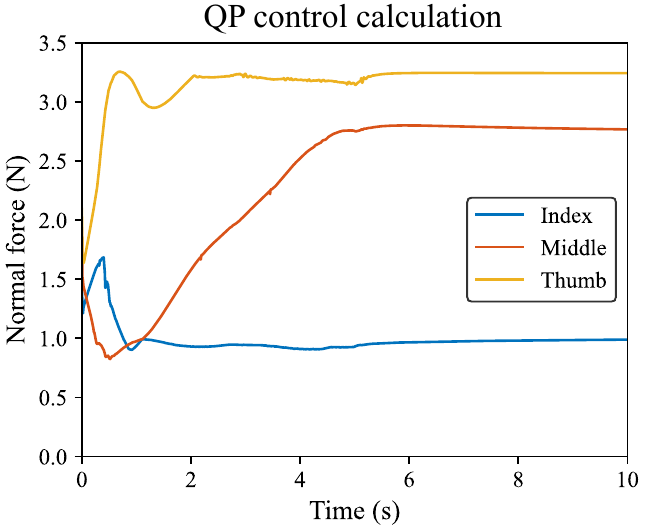}
    \caption{(Left) Fingertip normal forces for a controller based on the closed-form solution~\eqref{eq:sol_multi_hand}. The normal force at the middle finger drops to zero, and contact is lost. (Right) Fingertip normal forces for a controller based on the quadratic program~\eqref{eq:QP}. All fingertip normal forces are kept bounded away from zero. }
    \label{fig:qp-vs-cf}
\end{figure}

The advantage of the QP approach over the closed-form approach is demonstrated by the normal force plots of Figure~\ref{fig:qp-vs-cf} for the task of twisting a constrained cylinder. For each plot, the hand starts from the same initial grasp. Initially, the two controllers behave similarly. As fingertip force magnitudes drop, however, the behavior diverges, with the QP-based controller attempting to enforce a minimum total force magnitude of $1$~N at each finger. In the closed-form approach, which ignores force inequality constraints, the normal force of the middle finger drops to zero, contact is lost, and the manipulation fails. 

In a small number of simulated experiments, the closed-form controller was able to achieve the desired trajectory when the QP-based controller was unable, due to the fact that the closed-form controller has access to ``unsafe'' but potentially useful controls that the QP-based controller does not. These controls induce small contact forces or forces near the edges of friction cones.

\subsection{Hardware experiments} 
\label{ssec:experiments}

\begin{figure}
    \centering
    \includegraphics[width=\linewidth]{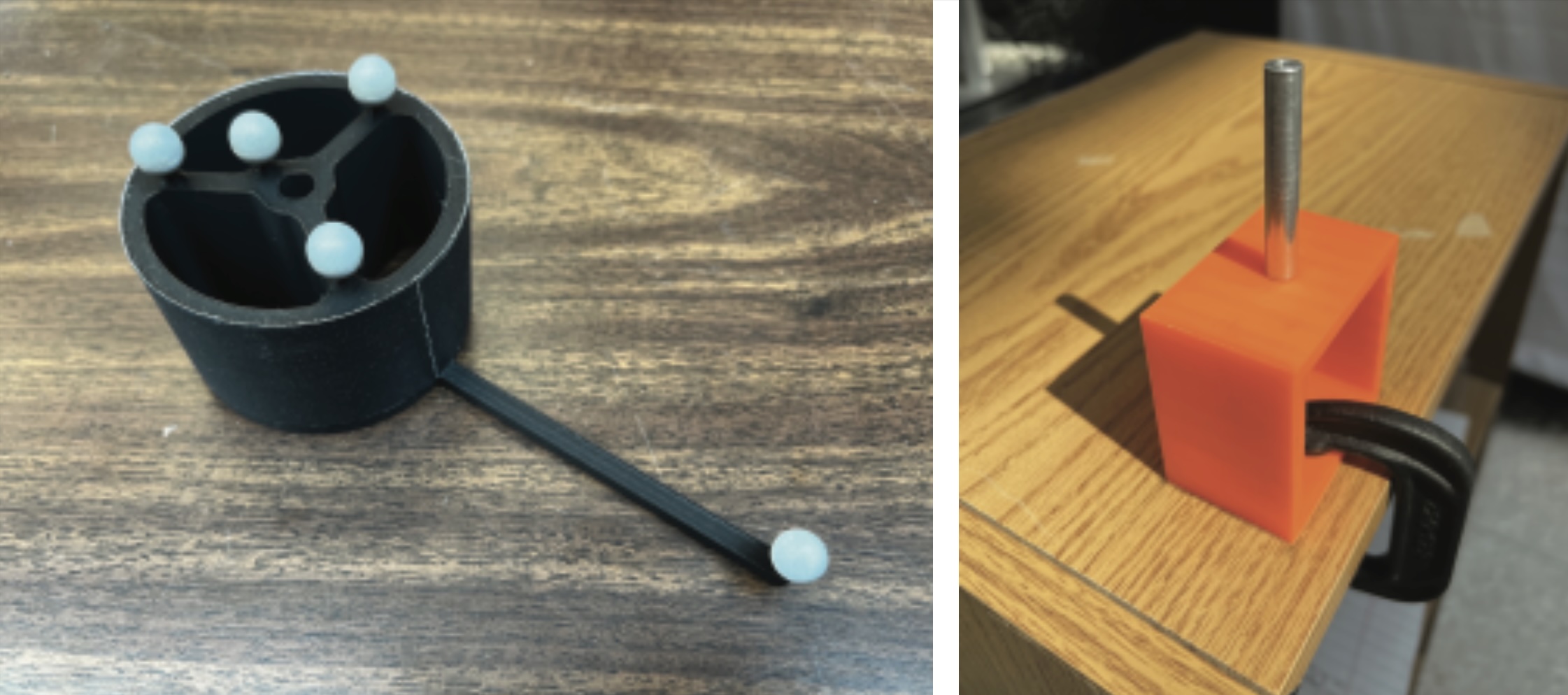}
    \caption{Left: A cylindrical object with a central axis. Right: An axis that allows only vertical and rotational motions of the cylindrical object.}
    \label{fig:manip-objects}
\end{figure}

Inspired by the lid-twisting example of Figure~\ref{fig:human_demo}, we tested in-hand rolling of the cylinder shown in Figure~\ref{fig:manip-objects}. The cylinder is constrained to rotate and translate along a vertical axis, and OptiTrack markers are used for motion tracking. The friction coefficient between the Visiflex fingertips and objects was estimated to be between 0.5 and 1.

The task was to use two fingers and the thumb of the Allegro hand to rotate the cylinder at a constant speed about its axis by $-30^\circ$ in $5$~s, and then to hold the cylinder at that orientation for another $5$~s (the $R_z$ trajectory). The fingers are initially brought into contact with the cylinder by a fixed motion of the WAM arm and Allegro hand. We rely on position control of the finger joints, along with the passive compliance of the fingertips, to form a stable grasp of the cylinder with initial fingertip normal forces of approximately 1 to 2~N, similar to the simulation study above. Then we engage the rolling controller of Figure~\ref{fig:new-control} using the closed-form solution~\eqref{eq:sol_multi_hand} for finger joint velocities. After the task is completed, the hand opens and the arm moves to a reset position. As in the simulation of constrained manipulation, the control gain associated with translational motion along the $z$-axis is zero.

\begin{figure}
\centering
\includegraphics[width=3.4in]{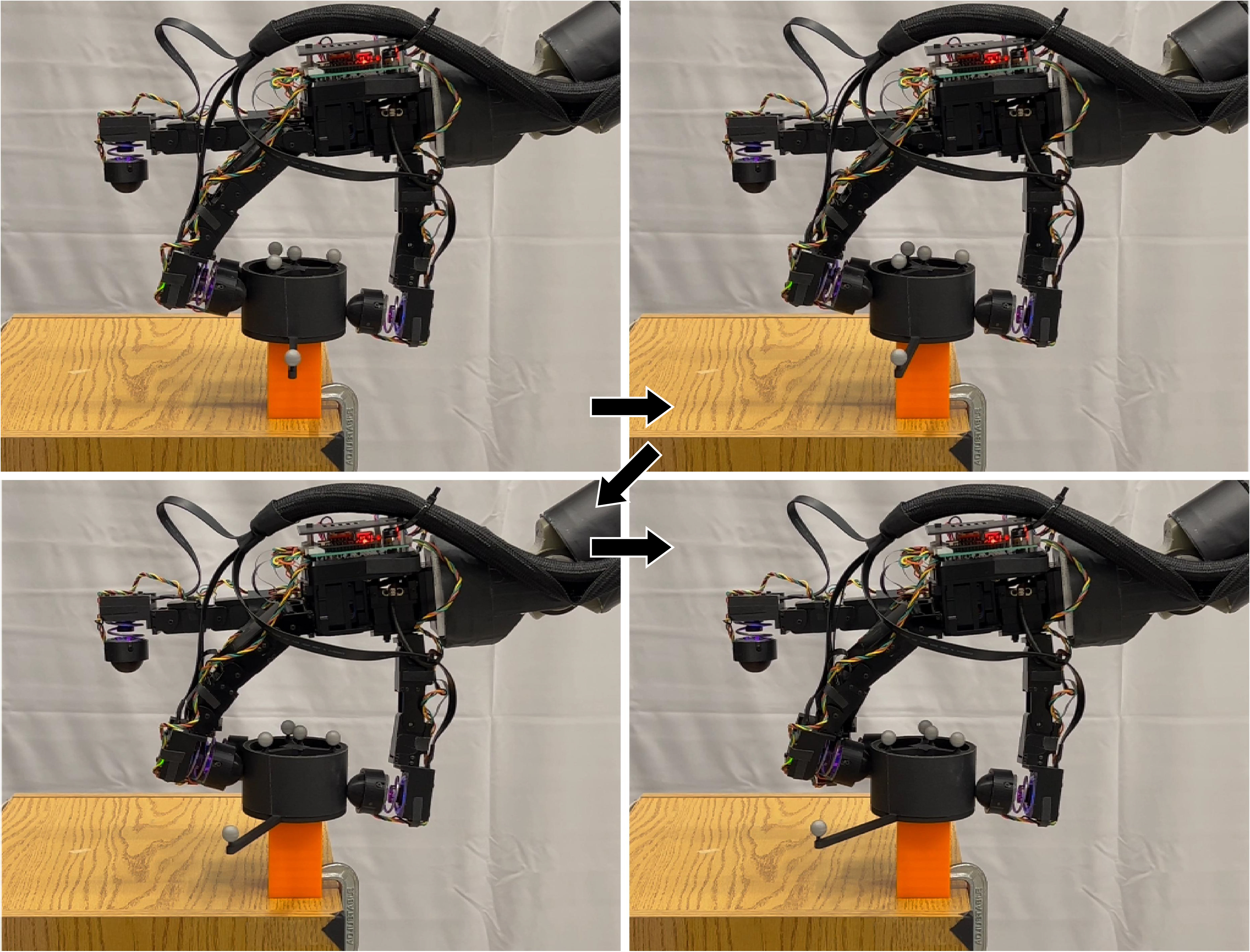}
\caption{Snapshots of in-hand rolling manipulation to rotate the cylinder.}
\label{fig:rotation_zoomin_sequence}
\end{figure}

\begin{figure}
\centering
\includegraphics[width=3.4in]{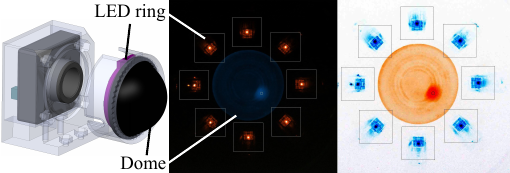}
\caption{(Left) Schematic of the Visiflex tactile sensor. (Middle) The embedded camera detects the contact location on the dome and the deflection of the eight LED fiducials, which allows calculation of the contact wrench due to the known flexure stiffness. (Right) A color-adjusted version of the camera's view.}
\label{fig:view_example}
\end{figure}

\begin{figure}
\centering
\includegraphics[width=3.4in]{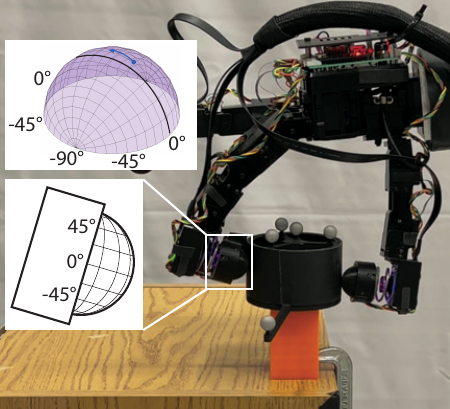}
\caption{The inset shows the evolution of the contact point on the robot's index fingertip during one execution of in-hand rolling manipulation.}
\label{fig:contact_trajectory}
\end{figure}

\begin{figure}
\centering
\includegraphics[width=3.4in]{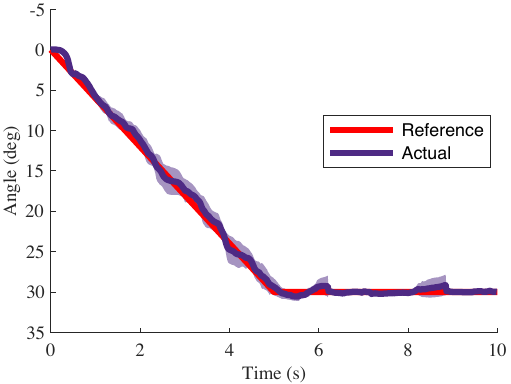}
\caption{Experimental results of $19$ runs of in-hand rotation of the cylinder, plotted as the mean and standard deviation. Trajectory tracking errors are small.}
\label{fig:rotating_ff_fb}
\end{figure}

\begin{figure}
\centering
\includegraphics[width=3.4in]{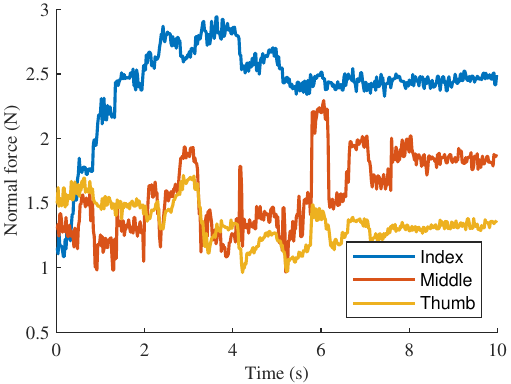}
\caption{The normal contact forces of all fingertips in an example run.}
\label{fig:normal_force_single_run}
\end{figure}

\begin{figure}
\centering
\includegraphics[width=3.4in]{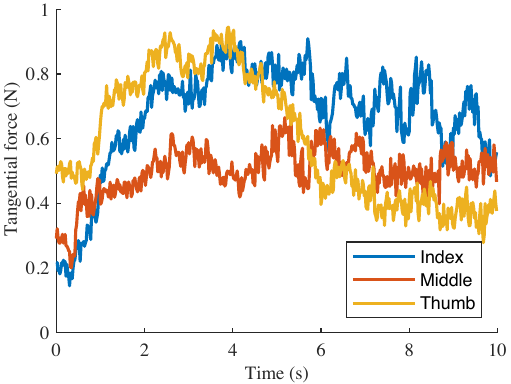}
\caption{The tangential contact forces of all fingertips in an example run.}
\label{fig:tangential_force_single_run}
\end{figure}

Figure~\ref{fig:rotation_zoomin_sequence} shows snapshots from one execution. Because each Allegro finger has only four joints, relative motion (e.g., rolling) is required at the fingertips to achieve rotation of the cylinder via in-hand manipulation. Figure~\ref{fig:view_example} shows the view from one of the Visiflex tactile sensors: the location of the object contact on the fingertip dome is tracked, and the motions of the eight LED fiducials are used to determine the flexure's displacement, and therefore the contact wrench from the known flexure stiffness. Figure~\ref{fig:contact_trajectory} shows the travel of the contact point on one of the fingertip domes due to rolling during one execution of the task. Figure~\ref{fig:rotating_ff_fb} shows the twisting angle results of 19 experimental runs. Experimental runs concluded after $10$~s or if a finger lost contact with the object. Figures~\ref{fig:normal_force_single_run} and \ref{fig:tangential_force_single_run} show the normal and tangential force plots of all three fingertips in an example run.  

As shown in Figure~\ref{fig:rotating_ff_fb}, the cylinder was successfully rotated to the desired angle at the desired velocity, and the tracking error was small. In our implementation, however, we used the closed-form solution~\eqref{eq:sol_multi_hand} to calculate commanded finger joint velocities instead of the quadratic program~\eqref{eq:QP}. Sufficient normal forces during manipulation were due to the residual of the initial pre-programmed grasp, not explicit enforcement of contact inequality constraints throughout the manipulation.  
As a result of not explicitly enforcing force inequality constraints, near the end of several of the experimental runs, one or more fingers broke contact with the cylinder.

%%%%%%%%%%%%%%%%%%%%%%%%%%%%%%%%%
%%%%%%%%%%%%%%%%%%%%%%%%%%%%%%%%%

\section{Conclusion} \label{cha:conclusion}

The hardware of robotic hands and tactile sensors will continue to evolve rapidly, as will AI and control of dexterity, but control of compliant hand rolling manipulation, as illustrated in Figures~\ref{fig:human_demo} and \ref{fig:chopsticks-3-types}, will remain a core dexterous capability, and the mechanics governing this interaction will remain unchanged. 

This paper derives the mechanics of quasistatic, compliant, in-hand rolling manipulation, including forward and inverse mechanics; proposes a feedback controller based on vision and tactile feedback; provides a simulation validation of the approach; and reports a preliminary experimental validation on the task of twisting a cylinder using three fingers of an Allegro hand outfitted with Visiflex tactile fingertips. 

This results of this paper are easily adapted to cases where the object interacts with an environment including springs and dampers, in addition to gravity, by modifying~\eqref{eq:F_ext}. If the object interacts with $k$ hard kinematic constraints (e.g., $k=3$ if the object is constrained to slide on a surface, or $k=4$ if it is constrained to translate and rotate about an axis), the inverse mechanics could be modified to allow the user to specify the rate of change of $k$ components of the wrench the object applies to the environment and an orthogonal $6-k$ components of the object's twist $\twist{}{wo}$. Contact with the environment may also be used to help achieve in-hand manipulation (e.g.,~\cite{dafle2014extrinsic,shi2020hand}).

The approach in this paper can be extended to manipulation of articulated rigid bodies, like scissors or a pipette, or to ``small'' soft contact patches by adding a torsional friction term in Section~\ref{subsec:force_equ}. In general, torsional friction tends to counteract spinning about the contact normal while allowing ``pure rolling''~\cite{Woodruff2023}. A different approach is required, however, for large contact patches, contact patches spanning multiple phalanges of the fingers, and deformable objects. 

In Section~\ref{ssec:preliminaries}, the assumption of small flexure displacements yields a linear relationship between the fingertip wrench and the exponential coordinates of the flexure displacement. This simplifies the form of the compliant rolling mechanics equations. For the small displacements observed in our experiments (orientation displacements of less than $20^\circ$ and linear displacements less than $10\%$ of the flexure's length), this linear approximation yields comparable wrench estimates to nonlinear fits to the data (less than $5\%$ error)~\cite{fernandez2021visiflex}.  For larger flexure displacements where a linear fit deteriorates, a nonlinear wrench-displacement relationship can be linearized at the current flexure configuration.

This paper's focus is the differential rolling motion within a grasp, not establishing the grasp in the first place. Our work could be augmented by existing approaches to grasp planning and, for large in-hand motions, finite in-hand rolling trajectory planning and finger gaiting, where one or more fingers detach and establish a new grasp before continuing with in-hand rolling manipulation. To implement finger gaiting, the hand must have enough fingers to hold the object in wrench equilibrium while one or more fingers are out of contact, and the finger joint velocity calculation~\eqref{eq:QP} must be modified to drive the contact force of a detaching finger to zero to allow it to break contact.

Future work includes more extensive experimental validation and robustification using the Allegro and other robot hands (among the many currently being produced) that exhibit compliance that can be modeled. We are also investigating data-driven approaches to in-hand manipulation, based on behavior cloning and reinforcement learning, that integrate models as explored in this paper. For example, models can provide a performance baseline, potentially allowing learning to focus on accounting for residual physics that are not well modeled (e.g., deformable contacts).  Also, models can be used to focus real and simulated robot experiments on physically meaningful interactions, and to potentially extract more generalizable information from each experiment, with the goal of more efficient robot learning.

%%%%%%%%%%%%%%%%%%%%%%%%%%%%%%%%%
%%%%%%%%%%%%%%%%%%%%%%%%%%%%%%%%%
%\clearpage

\appendix \label{cha:derivations}
\subsection{Derivation of Equation~\eqref{eq:dt_adj}} \label{cha:derivation_dt_adj}
\begin{equation*}
\begin{split}
&\frac{d}{dt}(\AdT{\text{ij}})\\
=&\begin{bmatrix}
\dot{\bf R}_\text{ij} & {\bf 0}_{3\times3}\\
\frac{d}{dt}([\p{i}{ij}]\R{\text{ij}}) & \dot{\bf R}_\text{ij}
\end{bmatrix}\\
=&\begin{bmatrix}
\dot{\bf R}^\intercal_\text{ji} & {\bf 0}_{3\times3}\\
\frac{d}{dt}([-\R{\text{ji}}^\intercal\p{j}{ji}]\R{\text{ij}}) & \dot{\bf R}^\intercal_\text{ji}
\end{bmatrix}~\text{(Using~\eqref{eq:p_rep})}\\
=&\begin{bmatrix}
\dot{\bf R}^\intercal_\text{ji} & {\bf 0}_{3\times3}\\
-\frac{d}{dt}(\R{\text{ji}}^\intercal[\p{j}{ji}]) & \dot{\bf R}^\intercal_\text{ji}
\end{bmatrix}\\
=&\begin{bmatrix}
\dot{\bf R}^\intercal_\text{ji} & {\bf 0}_{3\times3}\\
-\dot{\bf R}^\intercal_\text{ji}[\p{j}{ji}]-\R{\text{ji}}^\intercal[\dot{\bf p}^\text{j}_\text{ji}] & \dot{\bf R}^\intercal_\text{ji}
\end{bmatrix}~\text{(Using~\eqref{eq:skew_Rp})}\\
=&\begin{bmatrix}
([\ww{j}{ji}]\R{\text{ji}})^\intercal & {\bf 0}_{3\times3}\\
-([\ww{j}{ji}]\R{\text{ji}})^\intercal[\p{j}{ji}]-\R{\text{ji}}^\intercal[[\ww{j}{ji}]\p{j}{ji}+\vv{j}{ji}] & ([\ww{j}{ji}]\R{\text{ji}})^\intercal
\end{bmatrix}\\
&\text{(Using~\eqref{eq:dT})}\\
=&\begin{bmatrix}
-\R{\text{ji}}^\intercal[\ww{j}{ji}] & {\bf 0}_{3\times3}\\
\R{\text{ji}}^\intercal[\ww{j}{ji}][\p{j}{ji}]-\R{\text{ji}}^\intercal[[\ww{j}{ji}]\p{j}{ji}]-\R{\text{ji}}^\intercal[\vv{j}{ji}] & -\R{\text{ji}}^\intercal[\ww{j}{ji}]
\end{bmatrix}\\
=&
\begin{bmatrix}
-\R{\text{ij}}[\ww{j}{ji}] & {\bf 0}_{3\times3}\\
\R{\text{ij}}[\p{j}{ji}][\ww{j}{ji}]-\R{\text{ij}}[\vv{j}{ji}] & -\R{\text{ij}}[\ww{j}{ji}]
\end{bmatrix}~\text{(Using~\eqref{eq:skew_skew})}\\
=&\begin{bmatrix}
-\R{\text{ij}}[\ww{j}{ji}] & {\bf 0}_{3\times3}\\
\R{\text{ij}}[-\R{\text{ji}}\p{i}{ij}][\ww{j}{ji}]-\R{\text{ij}}[\vv{j}{ji}] & -\R{\text{ij}}[\ww{j}{ji}]
\end{bmatrix}\\
&\text{(Using~\eqref{eq:p_rep})}\\
=&-\begin{bmatrix}
\R{\text{ij}}[\ww{j}{ji}] & {\bf 0}_{3\times3}\\
[\p{i}{ij}]\R{\text{ij}}[\ww{j}{ji}]+\R{\text{ij}}[\vv{j}{ji}] & \R{\text{ij}}[\ww{j}{ji}]
\end{bmatrix}~\text{(Using~\eqref{eq:skew_Rp})}\\
=&-\AdT{\text{ij}}
\begin{bmatrix}
[\ww{j}{ji}] & \bf{0}_{3\times3}\\
[\vv{j}{ji}] & [\ww{j}{ji}]
\end{bmatrix}.
\end{split}
\end{equation*}

\subsection{Derivation of Equation~\eqref{eq:dt_wrench}} \label{cha:derivation_dt_wrench}
\begin{equation*}
\begin{split}
\dot{\mathcal{F}}^{\text{i}}
=&\frac{d}{dt}(\AdT{\text{ji}}^\intercal\wrench{j}{})=\frac{d}{dt}(\AdT{\text{ji}})^\intercal\wrench{j}{}+\AdT{\text{ji}}^\intercal\dot{\mathcal F}^{\text{j}}\\
=&(-\AdT{\text{ji}}
\begin{bmatrix}
[\ww{i}{ij}] & \bf{0}_{3\times3}\\
[\vv{i}{ij}] & [\ww{i}{ij}]
\end{bmatrix})^\intercal
\wrench{j}{}+\AdT{\text{ji}}^\intercal\dot{\mathcal F}^{\text{j}}\\
&\text{(Using~\eqref{eq:dt_adj})}\\
=&\begin{bmatrix}
[\ww{i}{ij}] & [\vv{i}{ij}]\\
\bf{0}_{3\times3} & [\ww{i}{ij}]
\end{bmatrix}
\AdT{\text{ji}}^\intercal\wrench{j}{}+\AdT{\text{ji}}^\intercal\dot{\mathcal{F}}^{\text{j}}\\
=&\begin{bmatrix}
[\ww{i}{ij}] & [\vv{i}{ij}]\\
\bf{0}_{3\times3} & [\ww{i}{ij}]
\end{bmatrix}
\wrench{i}{}+\AdT{\text{ji}}^\intercal\dot{\mathcal{F}}^{\text{j}}\\
=&-\W{i}{}\twist{i}{ij}+\AdT{\text{ji}}^\intercal\dot{\mathcal{F}}^{\text{j}}~\text{(Using~\eqref{eq:4skews})}.
\end{split}
\end{equation*}

\subsection{Derivation of Equation~\eqref{eq:object_final}} \label{cha:derivation_object_final}
\begin{equation*}
\begin{split}
\sum_{i=1}^{n}\dot{\mathcal F}_{\text{con},i}=&\sum_{i=1}^{n}(-{\bf W}_{\text{con},i}\twist{}{wc$_1,i$}+\AdT{\text{c$_1$w},i}^\intercal\dot{\mathcal{F}}^{\text{c}_1}_{\text{con},i})\\
&\text{(Using~\eqref{eq:dt_wrench})}\\
=&\sum_{i=1}^{n}(-{\bf W}_{\text{con},i}\twist{}{wc$_1,i$}+({\bf K}_{\text{spr},i}-{\bf W}_{\text{con},i})\twist{}{wa,$i$}\\
&~~~~~~-{\bf K}_{\text{spr},i}\twist{}{wf,$i$}+{\bf W}_{\text{con},i}(\twist{}{wo}+\twist{}{oc$_1,i$}))\\
&\text{(Using~\eqref{eq:df_con})}\\
=&\sum_{i=1}^{n}(({\bf K}_{\text{spr},i}-{\bf W}_{\text{con},i})\twist{}{wa,$i$}-{\bf K}_{\text{spr},i}\twist{}{wf,$i$}).
\end{split}
\end{equation*}
Taking the derivative of~\eqref{eq:wrench_equilib} and substituting this yields~\eqref{eq:object_final}.

\bibliographystyle{IEEEtran} %or another suitable style.
\bibliography{bib.bib}

@misc{shilati2026benchmarking,
      title={Benchmarking Dexterity of Multifingered Robot Hands: A Review and Perspective}, 
      author={Anthony Shilati and Anunth Ramaswami and Luke Batteas and Sylvia Tan and Anthony Barcio and Sairam Umakanth and Preksha Rao and David McDougall and Landry Graves and Ahmet A. Ozkan and Yunsoo Yoon and Arushi Pradhan and Michael G. Henry and Rohan Kota and Damian Gonzalez and Gray C. Thomas and Gary K. Fedder and J. Edward Colgate and Kevin M. Lynch},
      year={2026},
      eprint={2609.05585},
      archivePrefix={arXiv},
      primaryClass={cs.RO},
      url={https://arxiv.org/abs/2609.05585}, 
      note = "To appear, \textit{Annual Review of Control, Robotics, and Autonomous Systems}, 2027"
}

@article{goldberg2025gofe,
  author = {Goldberg, Ken},
  title = {Good old-fashioned engineering can close the 100,000-year ``data gap'' in robotics},
  journal = {Science Robotics},
  volume = {10},
  number = {105},
  pages = {eaea7390},
  year = {2025},
  publisher = {Science Robotics}
}

@misc{brooks2025dexterity,
  author = {Brooks, Rodney},
  title = {Why Today’s Humanoids Won’t Learn Dexterity},
  howpublished = {\url{https://rodneybrooks.com/why-todays-humanoids-wont-learn-dexterity/}},
  year = {2025},
  note = {Accessed: 2025-11-24}
}

@article{yang2025differentiable,
  title={Differentiable physics-based system identification for robotic manipulation of elastoplastic materials},
  author={Yang, Xintong and Ji, Ze and Lai, Yu-Kun},
  journal={The International Journal of Robotics Research},
  volume={44},
  number={13},
  pages={2126--2155},
  year={2025},
  publisher={SAGE Publications}
}

@inproceedings{si2024difftactile,
  title={{DIFFTACTILE}: A Physics-based Differentiable Tactile Simulator for Contact-rich Robotic Manipulation},
  author={Si, Zilin and Zhang, Gu and Ben, Qingwei and Romero, Branden and Xian, Zhou and Liu, Chao and Gan, Chuang},
  booktitle={International Conference on Learning Representations (ICLR)},
  year={2024}
}

@inproceedings{chen2025inhand,
  title={In-Hand Manipulation with Enforced Grasp Stability for Contact-Rich Tasks},
  author={Chen, Yifei and Lu, Shihan and Zhang, Haoxuan and Lynch, Kevin M.},
  booktitle={ICRA 2025 Workshop on Contact-Rich Manipulation},
  year={2025}
}

@inproceedings{zurbrugg2025graspqp,
  title={GraspQP: Differentiable Optimization of Force Closure for Diverse and Robust Dexterous Grasping},
  author={Zurbr{\"u}gg, Ren{\'e} and Cramariuc, Andrei and Hutter, Marco},
  booktitle={9th Conference on Robot Learning (CoRL)},
  year={2025}
}

@misc{yang2025cbfrl,
  title={CBF-RL: Safety Filtering Reinforcement Learning in Training with Control Barrier Functions},
  author={Yang, Lizhi and Werner, Blake and de Sa, Massimiliano and Ames, Aaron D.},
  year={2025},
  eprint={2510.14959},
  archivePrefix={arXiv},
  primaryClass={cs.RO}
}

@inproceedings{shang2025roboscape,
  title={RoboScape: Physics-informed Embodied World Model},
  author={Shang, Yu and Zhang, Xin and Tang, Yinzhou and Jin, Lei and Gao, Chen and Wu, Wei and Li, Yong},
  booktitle={Advances in Neural Information Processing Systems (NeurIPS)},
  year={2025}
}

@inproceedings{li2025pinwm,
  title={{PIN-WM}: Learning Physics-Informed World Models for Non-Prehensile Manipulation},
  author={Li, Wenxuan and Zhao, Hang and Yu, Zhiyuan and Du, Yu and Zou, Qin and Hu, Ruizhen and Xu, Kai},
  booktitle={Robotics: Science and Systems (RSS)},
  year={2025}
}

@article{zhu2020overview,
  title={The ingredients of real-world robotic reinforcement learning},
  author={Zhu, Henry and Yu, Justin and Gupta, Abhishek and Shah, Dhruv and Hartikainen, Kristian and Singh, Avi and Kumar, Vikash and Levine, Sergey},
  journal={arXiv preprint arXiv:2004.12570},
  year={2020}
}

@misc{WSJ2025,
title = {The `Hands Problem' Holding Back the Humanoid Revolution},
howpublished = {Wall Street Journal},
author = {Keilman, John},
month = oct,
year = 2025,
url = {https://www.wsj.com/tech/the-hands-problem-holding-back-the-humanoid-revolution-c1aa6123}}

@misc{HumDex2025,
title = {Dexterity for Humanoids Webinar Panel},
month = may,
year = 2025,
url = {https://www.youtube.com/watch?v=dB1mdFhO4oI}}

@ARTICLE{Woodruff2023,
  author={Woodruff, James Zachary and Lynch, Kevin M.},
  journal={IEEE Transactions on Robotics}, 
  title={Robotic Contact Juggling}, 
  year={2023},
  volume={39},
  number={3},
  pages={1964-1981}}

@inproceedings{baker1985stable,
  title={Stable prehension with a multi-fingered hand},
  author={Baker, B and Fortune, Steven and Grosse, Eric},
  booktitle={Proceedings. 1985 IEEE International Conference on Robotics and Automation},
  volume={2},
  pages={570--575},
  year={1985},
  organization={IEEE}
}

@article{sundaralingam2021hand,
  title={In-hand object-dynamics inference using tactile fingertips},
  author={Sundaralingam, Balakumar and Hermans, Tucker},
  journal={IEEE Transactions on Robotics},
  year={2021},
  publisher={IEEE}
}

@article{james2020slip,
  title={Slip Detection for Grasp Stabilization With a Multifingered Tactile Robot Hand},
  author={James, Jasper Wollaston and Lepora, Nathan F},
  journal={IEEE Transactions on Robotics},
  volume={37},
  number={2},
  pages={506--519},
  year={2020},
  publisher={IEEE}
}

@article{li2020review,
  title={A review of tactile information: Perception and action through touch},
  author={Li, Qiang and Kroemer, Oliver and Su, Zhe and Veiga, Filipe Fernandes and Kaboli, Mohsen and Ritter, Helge Joachim},
  journal={IEEE Transactions on Robotics},
  volume={36},
  number={6},
  pages={1619--1634},
  year={2020},
  publisher={IEEE}
}

@article{gualtieri2020learning,
  title={Learning manipulation skills via hierarchical spatial attention},
  author={Gualtieri, Marcus and Platt, Robert},
  journal={IEEE Transactions on Robotics},
  volume={36},
  number={4},
  pages={1067--1078},
  year={2020},
  publisher={IEEE}
}

@article{andrychowicz2020learning,
  title={Learning dexterous in-hand manipulation},
  author={Andrychowicz, OpenAI: Marcin and Baker, Bowen and Chociej, Maciek and Jozefowicz, Rafal and McGrew, Bob and Pachocki, Jakub and Petron, Arthur and Plappert, Matthias and Powell, Glenn and Ray, Alex and others},
  journal={The International Journal of Robotics Research},
  volume={39},
  number={1},
  pages={3--20},
  year={2020},
  publisher={SAGE Publications Sage UK: London, England}
}

@incollection{hou2020robust,
  title={Robust planar dynamic pivoting by regulating inertial and grip forces},
  author={Hou, Yifan and Jia, Zhenzhong and Johnson, Aaron M and Mason, Matthew T},
  booktitle={Algorithmic Foundations of Robotics XII},
  pages={464--479},
  year={2020},
  publisher={Springer}
}

@article{spiers2018variable,
  title={Variable-friction finger surfaces to enable within-hand manipulation via gripping and sliding},
  author={Spiers, Adam J and Calli, Berk and Dollar, Aaron M},
  journal={IEEE Robotics and Automation Letters},
  volume={3},
  number={4},
  pages={4116--4123},
  year={2018},
  publisher={IEEE}
}

@article{shi2020hand,
  title={In-hand sliding regrasp with spring-sliding compliance and external constraints},
  author={Shi, Jian and Weng, Huan and Lynch, Kevin M},
  journal={IEEE Access},
  volume={8},
  pages={88729--88744},
  year={2020},
  publisher={IEEE}
}

@article{costanzo2019two,
  title={Two-fingered in-hand object handling based on force/tactile feedback},
  author={Costanzo, Marco and De Maria, Giuseppe and Natale, Ciro},
  journal={IEEE Transactions on Robotics},
  volume={36},
  number={1},
  pages={157--173},
  year={2019},
  publisher={IEEE}
}

@article{hanafusa1977stable,
  title={Stable prehension of objects by the robot hand with elastic fingers},
  author={Hanafusa, Hideo and Asada, Haruhiko},
  journal={Transactions of the Society of Instrument and Control Engineers},
  volume={13},
  number={4},
  pages={370--377},
  year={1977},
  publisher={The Society of Instrument and Control Engineers}
}

@article{montana1988kinematics,
  title={The kinematics of contact and grasp},
  author={Montana, David J},
  journal={The International Journal of Robotics Research},
  volume={7},
  number={3},
  pages={17--32},
  year={1988},
  publisher={Sage Publications Sage CA: Thousand Oaks, CA}
}

@article{yuan2017gelsight,
  title={Gelsight: High-resolution robot tactile sensors for estimating geometry and force},
  author={Yuan, Wenzhen and Dong, Siyuan and Adelson, Edward H},
  journal={Sensors},
  volume={17},
  number={12},
  pages={2762},
  year={2017},
  publisher={Multidisciplinary Digital Publishing Institute}
}

@article{cutkosky1989computing,
  title={Computing and controlling compliance of a robotic hand},
  author={Cutkosky, Mark R and Kao, Imin},
  journal={IEEE transactions on robotics and automation},
  volume={5},
  number={2},
  pages={151--165},
  year={1989},
  publisher={IEEE}
}

@article{chavan2020planar,
  title={Planar in-hand manipulation via motion cones},
  author={Chavan-Dafle, Nikhil and Holladay, Rachel and Rodriguez, Alberto},
  journal={The International Journal of Robotics Research},
  volume={39},
  number={2-3},
  pages={163--182},
  year={2020},
  publisher={SAGE Publications Sage UK: London, England}
}

@inproceedings{dafle2014extrinsic,
  title={Extrinsic dexterity: In-hand manipulation with external forces},
  author={Dafle, Nikhil Chavan and Rodriguez, Alberto and Paolini, Robert and Tang, Bowei and Srinivasa, Siddhartha S and Erdmann, Michael and Mason, Matthew T and Lundberg, Ivan and Staab, Harald and Fuhlbrigge, Thomas},
  booktitle={2014 IEEE International Conference on Robotics and Automation (ICRA)},
  pages={1578--1585},
  year={2014},
  organization={IEEE}
}

@inproceedings{chen2015adaptive,
  title={An adaptive compliant multi-finger approach-to-grasp strategy for objects with position uncertainties},
  author={Chen, Zhaopeng and Wimb{\"o}ck, Thomas and Roa, Maximo A and Pleintinger, Benedikt and Neves, Miguel and Ott, Christian and Borst, Christoph and Lii, Neal Y},
  booktitle={2015 IEEE International Conference on Robotics and Automation (ICRA)},
  pages={4911--4918},
  year={2015},
  organization={IEEE}
}

@article{pozzi2016grasp,
  title={On grasp quality measures: Grasp robustness and contact force distribution in underactuated and compliant robotic hands},
  author={Pozzi, Maria and Malvezzi, Monica and Prattichizzo, Domenico},
  journal={IEEE Robotics and Automation Letters},
  volume={2},
  number={1},
  pages={329--336},
  year={2016},
  publisher={IEEE}
}

@article{dollar2010contact,
  title={Contact sensing and grasping performance of compliant hands},
  author={Dollar, Aaron M and Jentoft, Leif P and Gao, Jason H and Howe, Robert D},
  journal={Autonomous Robots},
  volume={28},
  number={1},
  pages={65--75},
  year={2010},
  publisher={Springer}
}

@inproceedings{bruyninckx1998generalized,
  title={Generalized stability of compliant grasps},
  author={Bruyninckx, Herman and Demey, Sabine and Kumar, Vijay},
  booktitle={Proceedings. 1998 IEEE International Conference on Robotics and Automation (Cat. No. 98CH36146)},
  volume={3},
  pages={2396--2402},
  year={1998},
  organization={IEEE}
}

@article{rodriguez2021unstable,
  title={The unstable queen: Uncertainty, mechanics, and tactile feedback},
  author={Rodriguez, Alberto},
  journal={Science Robotics},
  volume={6},
  number={54},
  pages={eabi4667},
  year={2021},
  publisher={American Association for the Advancement of Science}
}

@inproceedings{lynch1992manipulation,
  title={Manipulation and active sensing by pushing using tactile feedback.},
  author={Lynch, Kevin M and Maekawa, Hitoshi and Tanie, Kazuo},
  booktitle={IROS},
  volume={1},
  pages={416--421},
  year={1992}
}

@inproceedings{hogan2018tactile,
  title={Tactile regrasp: Grasp adjustments via simulated tactile transformations},
  author={Hogan, Francois R and Bauza, Maria and Canal, Oleguer and Donlon, Elliott and Rodriguez, Alberto},
  booktitle={2018 IEEE/RSJ International Conference on Intelligent Robots and Systems (IROS)},
  pages={2963--2970},
  year={2018},
  organization={IEEE}
}

@inproceedings{malvezzi2013evaluation,
  title={Evaluation of grasp stiffness in underactuated compliant hands},
  author={Malvezzi, Monica and Prattichizzo, Domenico},
  booktitle={2013 IEEE International conference on robotics and automation},
  pages={2074--2079},
  year={2013},
  organization={IEEE}
}

@article{ward2017model,
  title={Model-free precise in-hand manipulation with a 3d-printed tactile gripper},
  author={Ward-Cherrier, Benjamin and Rojas, Nicolas and Lepora, Nathan F},
  journal={IEEE Robotics and Automation Letters},
  volume={2},
  number={4},
  pages={2056--2063},
  year={2017},
  publisher={IEEE}
}

@inproceedings{van2015learning,
  title={Learning robot in-hand manipulation with tactile features},
  author={Van Hoof, Herke and Hermans, Tucker and Neumann, Gerhard and Peters, Jan},
  booktitle={2015 IEEE-RAS 15th International Conference on Humanoid Robots (Humanoids)},
  pages={121--127},
  year={2015},
  organization={IEEE}
}

@article{tegin2005tactile,
  title={Tactile sensing in intelligent robotic manipulation--a review},
  author={Tegin, Johan and Wikander, Jan},
  journal={Industrial Robot: An International Journal},
  year={2005},
  publisher={Emerald Group Publishing Limited}
}

@article{howe1993tactile,
  title={Tactile sensing and control of robotic manipulation},
  author={Howe, Robert D},
  journal={Advanced Robotics},
  volume={8},
  number={3},
  pages={245--261},
  year={1993},
  publisher={Taylor \& Francis}
}

@inproceedings{li2014localization,
  title={Localization and manipulation of small parts using gelsight tactile sensing},
  author={Li, Rui and Platt, Robert and Yuan, Wenzhen and ten Pas, Andreas and Roscup, Nathan and Srinivasan, Mandayam A and Adelson, Edward},
  booktitle={2014 IEEE/RSJ International Conference on Intelligent Robots and Systems},
  pages={3988--3993},
  year={2014},
  organization={IEEE}
}

@inproceedings{cole1988kinematics,
  title={Kinematics and control of multifingered hands with rolling contact},
  author={Cole, Arlene and Hauser, John and Sastry, Shankar},
  booktitle={Proceedings. 1988 IEEE International Conference on Robotics and Automation},
  pages={228--233},
  year={1988},
  organization={IEEE}
}

@article{cole1992dynamic,
  title={Dynamic control of sliding by robot hands for regrasping},
  author={Cole, Arlene A and Hsu, Ping and Sastry, S Shankar},
  journal={IEEE Transactions on robotics and automation},
  volume={8},
  number={1},
  pages={42--52},
  year={1992},
  publisher={IEEE}
}

@article{melchiorri2000slip,
  title={Slip detection and control using tactile and force sensors},
  author={Melchiorri, Claudio},
  journal={IEEE/ASME transactions on mechatronics},
  volume={5},
  number={3},
  pages={235--243},
  year={2000},
  publisher={IEEE}
}

@article{james2018slip,
  title={Slip detection with a biomimetic tactile sensor},
  author={James, Jasper Wollaston and Pestell, Nicholas and Lepora, Nathan F},
  journal={IEEE Robotics and Automation Letters},
  volume={3},
  number={4},
  pages={3340--3346},
  year={2018},
  publisher={IEEE}
}

@inproceedings{holweg1996slip,
  title={Slip detection by tactile sensors: algorithms and experimental results},
  author={Holweg, EGM and Hoeve, H and Jongkind, W and Marconi, Lorenzo and Melchiorri, Claudio and Bonivento, Claudio},
  booktitle={Proceedings of IEEE International Conference on Robotics and Automation},
  volume={4},
  pages={3234--3239},
  year={1996},
  organization={IEEE}
}

@inproceedings{ma2019dense,
  title={Dense tactile force estimation using GelSlim and inverse FEM},
  author={Ma, Daolin and Donlon, Elliott and Dong, Siyuan and Rodriguez, Alberto},
  booktitle={2019 International Conference on Robotics and Automation (ICRA)},
  pages={5418--5424},
  year={2019},
  organization={IEEE}
}

@article{shi2017dynamic,
  title={Dynamic in-hand sliding manipulation},
  author={Shi, Jian and Woodruff, J Zachary and Umbanhowar, Paul B and Lynch, Kevin M},
  journal={IEEE Transactions on Robotics},
  volume={33},
  number={4},
  pages={778--795},
  year={2017},
  publisher={IEEE}
}

@inproceedings{cirillo2017control,
  title={Control of linear and rotational slippage based on six-axis force/tactile sensor},
  author={Cirillo, Andrea and Cirillo, Pasquale and De Maria, Giuseppe and Natale, Ciro and Pirozzi, Salvatore},
  booktitle={2017 IEEE International Conference on Robotics and Automation (ICRA)},
  pages={1587--1594},
  year={2017},
  organization={IEEE}
}

@inproceedings{costanzo2018slipping,
  title={Slipping control algorithms for object manipulation with sensorized parallel grippers},
  author={Costanzo, Marco and De Maria, Giuseppe and Natale, Ciro},
  booktitle={2018 IEEE International Conference on Robotics and Automation (ICRA)},
  pages={7455--7461},
  year={2018},
  organization={IEEE}
}

@article{de2012force,
  title={Force/tactile sensor for robotic applications},
  author={De Maria, G and Natale, C and Pirozzi, S},
  journal={Sensors and Actuators A: Physical},
  volume={175},
  pages={60--72},
  year={2012},
  publisher={Elsevier}
}

@article{d2011silicone,
  title={Silicone-rubber-based tactile sensors for the measurement of normal and tangential components of the contact force},
  author={D'Amore, Alberto and De Maria, Giuseppe and Grassia, Luigi and Natale, Ciro and Pirozzi, Salvatore},
  journal={Journal of Applied Polymer Science},
  volume={122},
  number={6},
  pages={3757--3769},
  year={2011},
  publisher={Wiley Online Library}
}

@inproceedings{de2013slipping,
  title={Slipping control through tactile sensing feedback},
  author={De Maria, Giuseppe and Natale, Ciro and Pirozzi, Salvatore},
  booktitle={2013 IEEE International Conference on Robotics and Automation},
  pages={3523--3528},
  year={2013},
  organization={IEEE}
}

@inproceedings{donlon2018gelslim,
  title={Gelslim: A high-resolution, compact, robust, and calibrated tactile-sensing finger},
  author={Donlon, Elliott and Dong, Siyuan and Liu, Melody and Li, Jianhua and Adelson, Edward and Rodriguez, Alberto},
  booktitle={2018 IEEE/RSJ International Conference on Intelligent Robots and Systems (IROS)},
  pages={1927--1934},
  year={2018},
  organization={IEEE}
}

@article{ma2021extrinsic,
  title={Extrinsic Contact Sensing with Relative-Motion Tracking from Distributed Tactile Measurements},
  author={Ma, Daolin and Dong, Siyuan and Rodriguez, Alberto},
  journal={2021 IEEE International Conference on Robotics and Automation (ICRA)},
  pages={11262-11268},
  year={2021},
  organization={IEEE}
}

@book{lynch2017modern,
  title={Modern Robotics},
  author={Lynch, Kevin M and Park, Frank C},
  year={2017},
  publisher={Cambridge University Press}
}

@article{she2021cable,
  title={Cable manipulation with a tactile-reactive gripper},
  author={She, Yu and Wang, Shaoxiong and Dong, Siyuan and Sunil, Neha and Rodriguez, Alberto and Adelson, Edward},
  journal={The International Journal of Robotics Research},
  volume={40},
  number={12-14},
  pages={1385--1401},
  year={2021},
  publisher={SAGE Publications Sage UK: London, England}
}

@article{fernandez2021visiflex,
  title={Visiflex: A Low-Cost Compliant Tactile Fingertip for Force, Torque, and Contact Sensing},
  author={Fernandez, Alfonso J and Weng, Huan and Umbanhowar, Paul B and Lynch, Kevin M},
  journal={IEEE Robotics and Automation Letters},
  volume={6},
  number={2},
  pages={3009--3016},
  year={2021},
  publisher={IEEE}
}

@article{levine2016end,
  title={End-to-end training of deep visuomotor policies},
  author={Levine, Sergey and Finn, Chelsea and Darrell, Trevor and Abbeel, Pieter},
  journal={The Journal of Machine Learning Research},
  volume={17},
  number={1},
  pages={1334--1373},
  year={2016},
  publisher={JMLR. org}
}

@inproceedings{dong2019maintaining,
  title={Maintaining grasps within slipping bounds by monitoring incipient slip},
  author={Dong, Siyuan and Ma, Daolin and Donlon, Elliott and Rodriguez, Alberto},
  booktitle={2019 International Conference on Robotics and Automation (ICRA)},
  pages={3818--3824},
  year={2019},
  organization={IEEE}
}

@article{di2024using,
  title={Using fiber optic bundles to miniaturize vision-based tactile sensors},
  author={Di, Julia and Dugonjic, Zdravko and Fu, Will and Wu, Tingfan and Mercado, Romeo and Sawyer, Kevin and Most, Victoria Rose and Kammerer, Gregg and Speidel, Stefanie and Fan, Richard E and others},
  journal={IEEE Transactions on Robotics},
  year={2024},
  publisher={IEEE}
}

@article{zhao2024tac,
  title={Tac-man: Tactile-informed prior-free manipulation of articulated objects},
  author={Zhao, Zihang and Li, Yuyang and Li, Wanlin and Qi, Zhenghao and Ruan, Lecheng and Zhu, Yixin and Althoefer, Kaspar},
  journal={IEEE Transactions on Robotics},
  year={2024},
  publisher={IEEE}
}

@article{lee2025trajectory,
  title={Trajectory Optimization for In-Hand Manipulation with Tactile Force Control},
  author={Lee, Haegu and Kim, Yitaek and Staven, Victor Melbye and Sloth, Christoffer},
  journal={arXiv preprint arXiv:2503.08222},
  year={2025}
}

@article{suresh2024neuralfeels,
  title={NeuralFeels with neural fields: Visuotactile perception for in-hand manipulation},
  author={Suresh, Sudharshan and Qi, Haozhi and Wu, Tingfan and Fan, Taosha and Pineda, Luis and Lambeta, Mike and Malik, Jitendra and Kalakrishnan, Mrinal and Calandra, Roberto and Kaess, Michael and others},
  journal={Science Robotics},
  volume={9},
  number={96},
  pages={eadl0628},
  year={2024},
  publisher={American Association for the Advancement of Science}
}

@inproceedings{yuan2024robot,
  title={Robot synesthesia: In-hand manipulation with visuotactile sensing},
  author={Yuan, Ying and Che, Haichuan and Qin, Yuzhe and Huang, Binghao and Yin, Zhao-Heng and Lee, Kang-Won and Wu, Yi and Lim, Soo-Chul and Wang, Xiaolong},
  booktitle={2024 IEEE International Conference on Robotics and Automation (ICRA)},
  pages={6558--6565},
  year={2024},
  organization={IEEE}
}

@article{khadivar2023adaptive,
  title={Adaptive fingers coordination for robust grasp and in-hand manipulation under disturbances and unknown dynamics},
  author={Khadivar, Farshad and Billard, Aude},
  journal={IEEE Transactions on Robotics},
  volume={39},
  number={5},
  pages={3350--3367},
  year={2023},
  publisher={IEEE}
}

@inproceedings{rostel2023estimator,
  title={Estimator-coupled reinforcement learning for robust purely tactile in-hand manipulation},
  author={R{\"o}stel, Lennart and Pitz, Johannes and Sievers, Leon and B{\"a}uml, Berthold},
  booktitle={2023 IEEE-RAS 22nd International Conference on Humanoid Robots (Humanoids)},
  pages={1--8},
  year={2023},
  organization={IEEE}
}

@inproceedings{pitz2024learning,
  title={Learning time-optimal and speed-adjustable tactile in-hand manipulation},
  author={Pitz, Johannes and R{\"o}stel, Lennart and Sievers, Leon and B{\"a}uml, Berthold},
  booktitle={2024 IEEE-RAS 23rd International Conference on Humanoid Robots (Humanoids)},
  pages={973--979},
  year={2024},
  organization={IEEE}
}

@article{zhou2024dexterous,
  title={A dexterous and compliant (dexco) hand based on soft hydraulic actuation for human inspired fine in-hand manipulation},
  author={Zhou, Jianshu and Huang, Junda and Dou, Qi and Abeel, Pieter and Liu, Yunhui},
  journal={IEEE Transactions on Robotics},
  year={2024},
  publisher={IEEE}
}

@article{HU2025104904,
title = {Dexterous in-hand manipulation of slender cylindrical objects through deep reinforcement learning with tactile sensing},
journal = {Robotics and Autonomous Systems},
volume = {186},
pages = {104904},
year = {2025},
issn = {0921-8890},
doi = {https://doi.org/10.1016/j.robot.2024.104904},
url = {https://www.sciencedirect.com/science/article/pii/S0921889024002884},
author = {Wenbin Hu and Bidan Huang and Wang Wei Lee and Sicheng Yang and Yu Zheng and Zhibin Li},
}

@inproceedings{Roestel_2025,
   title={Composing Dextrous Grasping and In-hand Manipulation via Scoring with a Reinforcement Learning Critic},
   booktitle={2025 IEEE International Conference on Robotics and Automation (ICRA)},
   publisher={IEEE},
   author={Röstel, Lennart and Winkelbauer, Dominik and Pitz, Johannes Sievers, Leon and Bäuml, Berthold},
   year={2025},
   month=may }

@inproceedings{lee2025vitascope,
  title={ViTaSCOPE: Visuo-tactile Implicit Representation for In-hand Pose and Extrinsic Contact Estimation},
  author={Lee, Jayjun and Fazeli, Nima},
  booktitle={Robotics: Science and Systems (RSS)},
  year={2025}
}

@inproceedings{shirai2025pivoting,
  title={Sim-to-Real Contact-Rich Pivoting via Optimization-Guided RL with Vision and Touch},
  author={Shirai, Yuki and Ota, Kei and Jha, Devesh K. and Romeres, Diego},
  booktitle={NeurIPS Workshop on Embodied World Models},
  year={2025}
}

@misc{Abhishek2026residual,
      title={RFS: Reinforcement Learning with Residual Flow Steering for Dexterous Manipulation}, 
      author={Entong Su and Tyler Westenbroek and Anusha Nagabandi and Abhishek Gupta},
      year={2026},
      eprint={2602.01789},
      archivePrefix={arXiv},
      primaryClass={cs.RO},
      url={https://arxiv.org/abs/2602.01789}, 
}

@misc{gupta2025grasp,
  title={Grasp to Act: Dexterous Grasping for Tool Use in Dynamic Settings},
  author={Gupta, Harsh and Mirzaee, Mohammad Amin and Yuan, Wenzhen},
  year={2025},
  note={Available at \url{https://grasp2act.github.io/}}
}

@ARTICLE{rgmc2024geometryfree,
  author={Yu, Mingrui and Jiang, Yongpeng and Chen, Chen and Jia, Yongyi and Li, Xiang},
  journal={IEEE Robotics and Automation Letters}, 
  title={Robotic In-Hand Manipulation for Large-Range Precise Object Movement: The RGMC Champion Solution}, 
  year={2025},
  volume={10},
  number={5},
  pages={4738-4745},
  doi={10.1109/LRA.2025.3555138}}

@misc{hidex2025,
      title={Enhancing Dexterity in Robotic Manipulation via Hierarchical Contact Exploration}, 
      author={Xianyi Cheng and Sarvesh Patil and Zeynep Temel and Oliver Kroemer and Matthew T. Mason},
      year={2023},
      eprint={2307.00383},
      archivePrefix={arXiv},
      primaryClass={cs.RO},
      url={https://arxiv.org/abs/2307.00383}, 
}

@misc{liu2025dexndm,
      title={DexNDM: Closing the Reality Gap for Dexterous In-Hand Rotation via Joint-Wise Neural Dynamics Model}, 
      author={Xueyi Liu and He Wang and Li Yi},
      year={2025},
      eprint={2510.08556},
      archivePrefix={arXiv},
      primaryClass={cs.RO},
      url={https://arxiv.org/abs/2510.08556}, 
}

@INPROCEEDINGS{qi2025simplecomplexskillscase,
  author={Qi, Haozhi and Yi, Brent and Lambeta, Mike and Ma, Yi and Calandra, Roberto and Malik, Jitendra},
  booktitle={2025 IEEE International Conference on Robotics and Automation (ICRA)}, 
  title={From Simple to Complex Skills: The Case of In-Hand Object Reorientation}, 
  year={2025},
  volume={},
  number={},
  pages={14291-14298},
  doi={10.1109/ICRA55743.2025.11128016}}

\vspace*{-1in}
\begin{IEEEbiography}[{\includegraphics[width=1in,height=1.25in,clip,keepaspectratio]{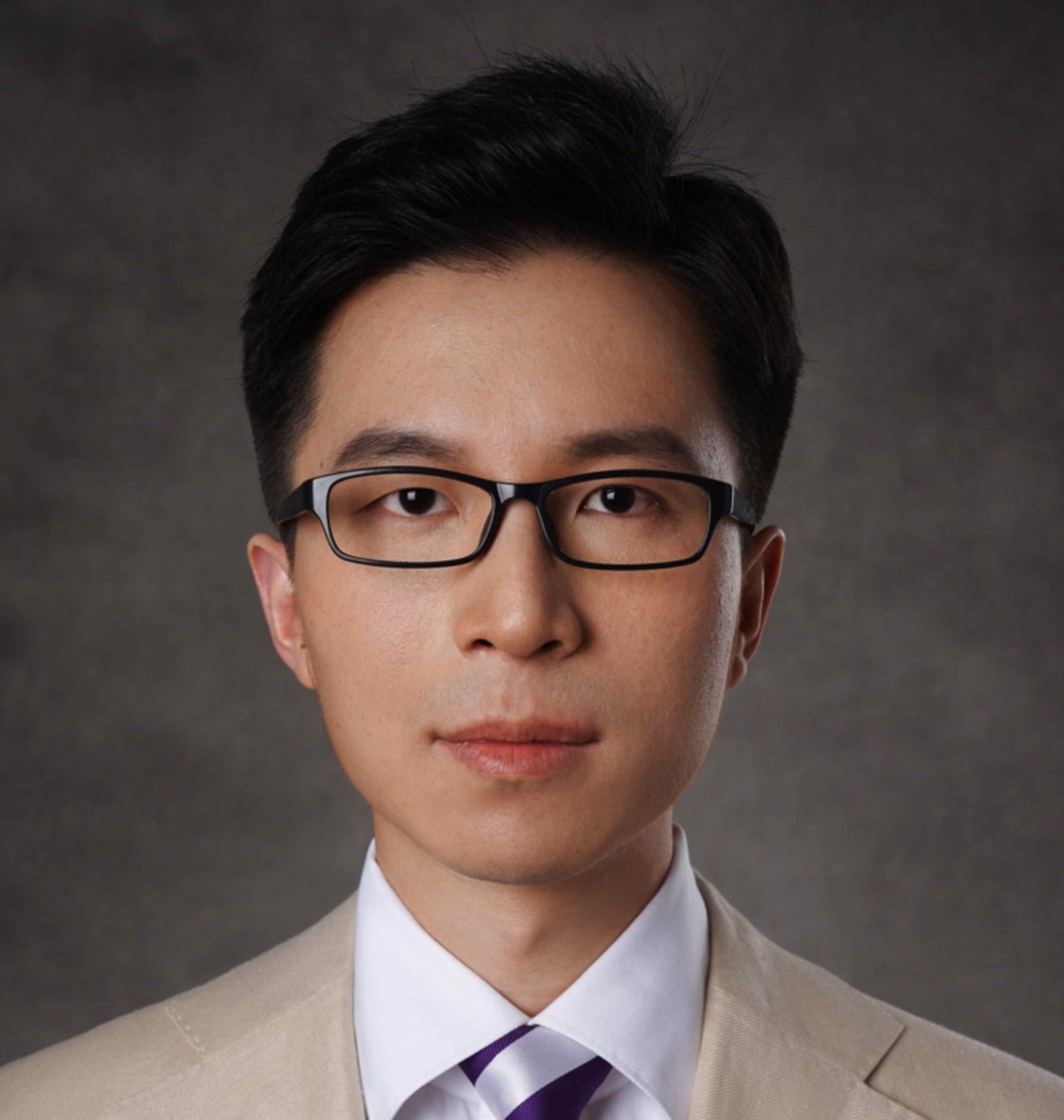}}]{Huan Weng} received the B.S. degree in mechanical engineering and automation from Beihang University, Beijing, China, in 2014, and the M.S. and Ph.D. degrees in mechanical engineering from Northwestern University, Evanston, IL, USA, in 2017 and 2024, respectively. At Northwestern University, he was affiliated with the Center for Robotics and Biosystems. He is currently an applied scientist at Sudo AI, Shanghai, China. His research interests include robotic manipulation and control.
\end{IEEEbiography}

\vspace*{-1in}
\begin{IEEEbiography}[{\includegraphics[width=1in,height=1.25in,clip,keepaspectratio]{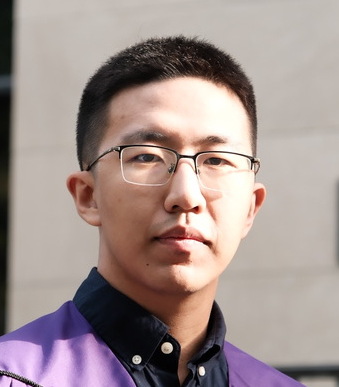}}]{Yifei Chen} received the B.S. degree in robotics engineering from the Southern University of Science and Technology, Shenzhen, China, in 2024, and the M.S. degree in mechanical engineering from Northwestern University, Evanston, IL, USA, in 2026, where he was affiliated with the Center for Robotics and Biosystems.

He is currently a Robotics Engineer. His research interests include dexterous robotic manipulation, hardware–control co-design, and reinforcement learning.
\end{IEEEbiography}

\vspace*{-1in}
\begin{IEEEbiography}[{\includegraphics[width=1in,height=1.25in,clip,keepaspectratio]{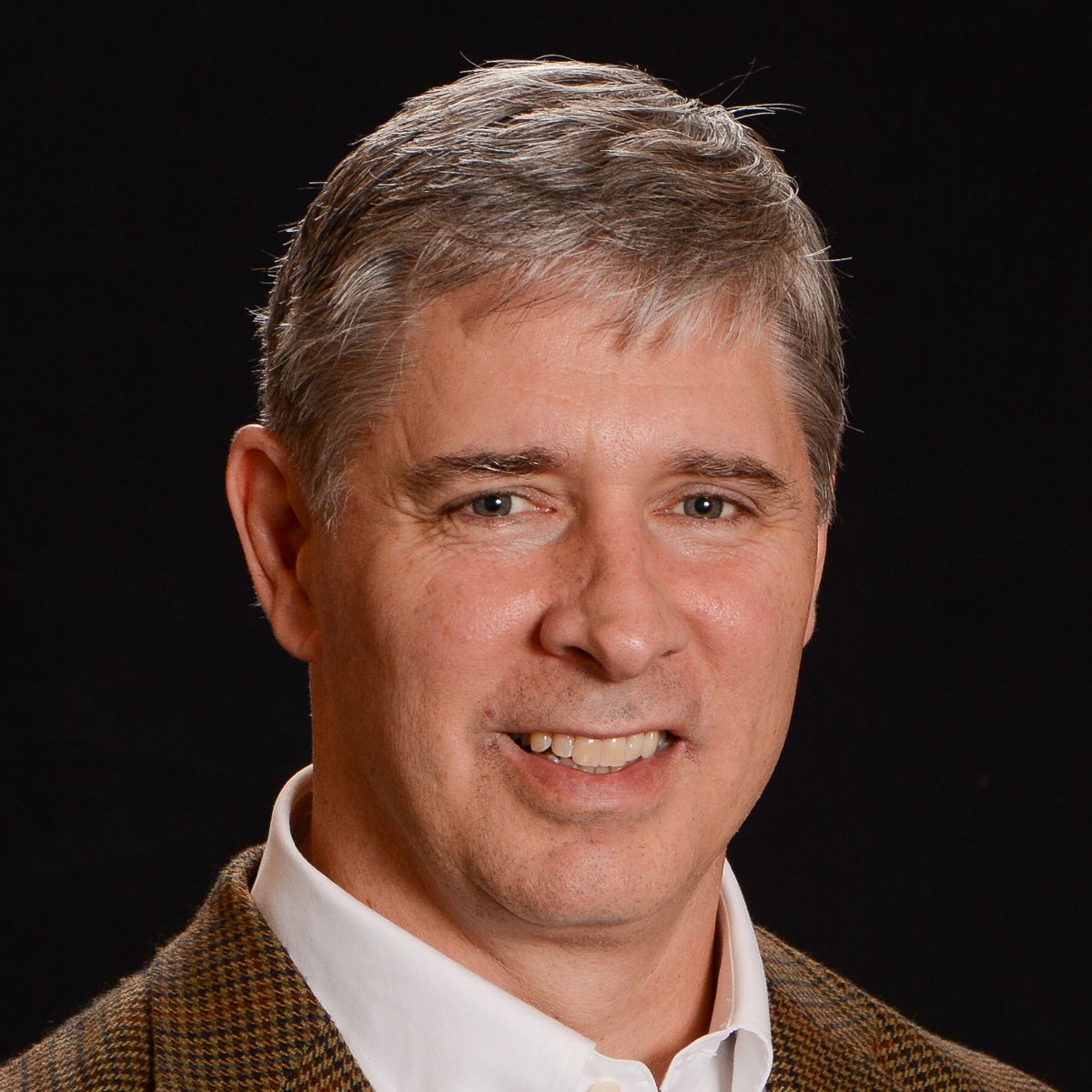}}]{Kevin M. Lynch} (Fellow, IEEE) received the B.S.E. degree in electrical engineering from Princeton University, Princeton, NJ, USA, and the Ph.D. degree in robotics from Carnegie Mellon University, Pittsburgh, PA, USA.

He is currently a Professor of Mechanical Engineering with Northwestern University, Chicago, IL, USA, where he directs the Center for Robotics and Biosystems and is research director of the US NSF HAND Engineering Research Center. He has coauthored the textbooks \textit{Principles of Robot Motion} (Cambridge, MA, USA: MIT Press, 2005) and \textit{Modern Robotics: Mechanics, Planning, and Control} (Cambridge, U.K.: Cambridge Univ. Press, 2017) and the associated online courses and videos. His research interests include robot manipulation and locomotion, self-organizing multiagent systems, and physically collaborative human–robot systems.
\end{IEEEbiography}

\end{document}